\documentclass[lettersize,journal, compsoc]{IEEEtran} 
\usepackage{amsmath,amsfonts}
\usepackage{algorithmic}
\usepackage{algorithm}
\usepackage{array}
\usepackage[caption=false,font=normalsize,labelfont=sf,textfont=sf]{subfig}
\usepackage{textcomp}
\usepackage{stfloats}
\usepackage{url}
\usepackage{verbatim}
\usepackage{graphicx}
\usepackage{cite}
\usepackage{ragged2e} 
\usepackage{cellspace}
\usepackage{xcolor}
\DeclareMathOperator*{\argmin}{arg\,min}

\begin{document}

\title{Learning Radiance Fields from a Single Snapshot Compressive Image}

\author{Yunhao Li\IEEEauthorrefmark{1*}, Xiang Liu\IEEEauthorrefmark{1*}, Xiaodong Wang, Xin Yuan\IEEEauthorrefmark{2} and Peidong Liu\IEEEauthorrefmark{2}
\thanks{Yunhao Li is with the College of Computer Science and Technology at Zhejiang University and the School of Engineering at Westlake University, Hangzhou, Zhejiang, China. (Email:\emph{liyunhao}@westlake.edu.cn).

All the other authors are with the School of Engineering at Westlake University, Hangzhou, Zhejiang, China. Email: ({\emph{liuxiang}, \emph{wangxiaodong}, \emph{xyuan}, \emph{liupeidong}}@westlake.edu.cn). }
\thanks{*Yunhao Li and Xiang Liu contributed equally.}
\thanks{$\dagger$Peidong Liu and Xin Yuan are the corresponding authors.}}

\markboth{Journal of \LaTeX\ Class Files,~Vol.~14, No.~8, August~2021}%
{Shell \MakeLowercase{\textit{et al.}}: A Sample Article Using IEEEtran.cls for IEEE Journals}


\IEEEtitleabstractindextext{ 
\begin{abstract}
\justifying
In this paper, we explore the potential of Snapshot Compressive Imaging (SCI) technique for recovering the underlying 3D scene structure from a single temporal compressed image. SCI is a cost-effective method that enables the recording of high-dimensional data, such as hyperspectral or temporal information, into a single image using low-cost 2D imaging sensors. To achieve this, a series of specially designed 2D masks are usually employed, reducing storage and transmission requirements and offering potential privacy protection. Inspired by this, we take one step further to recover the encoded 3D scene information leveraging powerful 3D scene representation capabilities of neural radiance fields (NeRF). Specifically, we propose SCINeRF, in which we formulate the physical imaging process of SCI as part of the training of NeRF, allowing us to exploit its impressive performance in capturing complex scene structures. In addition, we further integrate the popular 3D Gaussian Splatting (3DGS) framework and propose SCISplat to improve 3D scene reconstruction quality and training/rendering speed by explicitly optimizing point clouds into 3D Gaussian representations. To assess the effectiveness of our method, we conduct extensive evaluations using both synthetic data and real data captured by our SCI system. Experimental results demonstrate that our proposed approach surpasses the state-of-the-art methods in terms of image reconstruction and novel view synthesis. Moreover, our method also exhibits the ability to render high frame-rate multi-view consistent images in real time by leveraging SCI and the rendering capabilities of 3DGS.
Codes will be available at: \textcolor{magenta}{https://github.com/WU-CVGL/SCISplat}.


\end{abstract}
} 

\maketitle

\section{Introduction}
Capturing and reconstructing 3D scenes from a single image is challenging. To address these challenges, we present a practical approach to reconstruct the underlying 3D scene from a single snapshot compressive image. Our method is built upon the video SCI system \cite{yuan2021snapshot}, a novel approach which is originally designed to capture high-speed scenes~\cite{Llull13_OE_CACTI}. Conventional high-speed imaging systems often face challenges like high hardware costs and storage requirements. Drawing inspiration from Compressed Sensing (CS) \cite{Candes2006TIT, Donoho2006TIT}, video SCI system has emerged to mitigate these limitations. A typical video SCI system comprises a hardware encoder and a software decoder. The hardware encoder employs a series of specially designed (or random binary distributed) two-dimensional masks to modulate the incoming photons across exposure time into a single compressed image/measurement. This approach enables the low-cost CCD/CMOS cameras to capture high-speed scenes, thereby reducing the storage requirement. The whole encoding process can also be achieved via software implementation on pre-captured images, which further minimizes storage and transmission demands while also providing additional privacy protection. On the other hand, the software decoder restores the high frame-rate images using the compressed measurement and corresponding binary masks.

\begin{figure*}[!htbp]
  \centering
    \includegraphics[width=\linewidth]{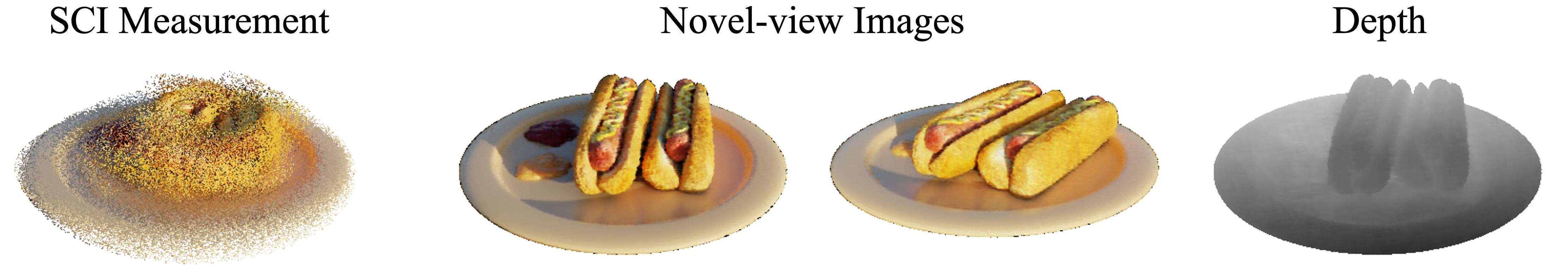}
    \caption{Given a single snapshot compressive image, our method can recover its underlying 3D scene representation. Leveraging the strong 3D scene representation and novel-view image synthesis capabilities of NeRF or 3DGS, we can recover high-quality multi-view consistent images from \textbf{the single measurement}.}
    \label{fig1}
\end{figure*}

In recent years, a variety of image reconstruction algorithms have been proposed for SCI, ranging from model-based methods \cite{yuan2015generalized, liao2014generalized, liu2018rank} to deep learning-based approaches \cite{qiao2020deep, cheng2020birnat, cheng2021memory, ma2019deep, wang2022spatial, wang2023efficientsci, cao2024hybrid}. These algorithms are capable of reconstructing encoded images or video frames with relatively high quality. However, these methods typically do not account for the underlying 3D scene structure necessary to ensure multi-view consistency and are limited to recovering images corresponding to the applied encoding masks. To address these limitations, we propose SCINeRF, the first end-to-end approach that recovers the underlying 3D scene representation from a single compressed image. The learned 3D scene representation enables the rendering of high-frame-rate, multi-view consistent images, as illustrated in Fig.~\ref{fig1}.

SCINeRF exploits neural radiance fields \cite{mildenhall2021nerf} to represent the 3D scene implicitly. Different from prior 3D representation techniques, NeRF takes pixel location and its corresponding camera pose as input and predicts the pixel intensity by volume rendering. This characteristic makes it well-suited for modeling the pixel-wise modulation process of an SCI system. Since it is impossible to recover the camera poses from a single compressed image via COLMAP \cite{schonberger2016structure}, we conduct a joint optimization on both the camera poses and NeRF by minimizing the difference between the synthesized compressed image and real measurement. Due to the fact that SCINeRF performs test-time optimization, it does not suffer from the generalization performance degradation as prior end-to-end deep learning-based methods. 
%
%
With SCINeRF, we can recover the 3D scene structure captured by a fast-moving camera with a short exposure time, {\em e.g.}, less than 20ms.

Nonetheless, due to the extensive implicit scene representation of NeRF and the time-consuming volumetric rendering process, SCINeRF suffers from a substantially long reconstruction time of about 5 hours for a single SCI image and a slow inference speed of 0.25FPS. A recent advancement, 3D Gaussian Splatting (3DGS)~\cite{kerbl20233d} extends the implicit neural rendering to explicit 3D Gaussians. By projecting these optimized Gaussians onto the image plane, 3DGS enables real-time rendering, which significantly improves both the efficiency and rendering quality compared to NeRF. Toward this end, we extend SCINeRF with 3DGS and propose {\bf SCISplat} to efficiently reconstruct 3D scene structure from an SCI measurement. Like SCINeRF, we cannot naively integrate 3DGS with SCI decoding because COLMAP cannot work with SCI images, while training 3DGS relies on point clouds and accurate camera poses for initialization. To address these challenges, we propose a novel initialization protocol, which first decodes a sequence of degraded frames with low-frequency cues from a single SCI image via pixel interpolation, then applies a learning-based structure from motion (SfM) model to robustly estimate the initial point cloud and camera poses from these degraded frames. Moreover, to better adapt 3DGS for ill-posed SCI decoding, we densify Gaussians with Monte Carlo Markov Chain (MCMC) \cite{kheradmand20243d} strategy that effectively suppresses noise. With SCISplat, we further improve the 3D scene reconstruction quality, and realize real-time high frame-rate image rendering from learned 3D representation.

To better evaluate the performance of our method, we set up \textbf{a real hardware platform} to collect real snapshot compressed images. We also conduct quantitative evaluations using synthetic images generated via Blender. Experimental results on both synthetic and real-world datasets demonstrate that SCINeRF and SCISplat achieve superior performance over previous state-of-the-art methods in terms of image restoration and novel view image synthesis. Regarding the efficiency, our SCISplat can consistently yield better results compared with state-of-the-art (SOTA) methods in terms of training and rendering/inference speed. In summary, our contributions are listed as follows:

\begin{itemize}
    \item[1)] We present the first method to restore 3D aware multi-view images from a single snapshot compressive image.
    \item[2)] We experimentally validate that our approach is able to synthesize high-quality novel view images, surpassing existing state-of-the-art SCI image/video reconstruction methods.
    \item[3)] Our methods also present an alternative approach for both 3D scene efficient storage/transmission and privacy protection transmission between the edge devices and cloud infrastructure for practical NeRF and 3DGS deployment.
\end{itemize}

The preliminary of this work was presented by authors and published on CVPR 2024 \cite{li2024scinerf}. In this  work, we extend the original work in the following significant ways:

\begin{itemize}
    \item[4)] We further enhance the scene reconstruction quality and training/rendering speed of SCINeRF by introducing SCISplat, a 3DGS-based approach, to reconstruct the underlying 3D scene from a single SCI image.
    \item[5)] We propose a novel initialization protocol to kick off 3DGS training by estimating point clouds and camera poses from a single SCI image robustly, which will benefit downstream 3D tasks from SCI measurements.
    \item[6)] All the experimental results related to SCISplat are newly conducted;
    \item[7)] Extensive experiments on both synthetic and real datasets show that SCISplat outperforms SCINeRF by \textbf{2.3dB} on image quality, \textbf{820$\times$} inference/rendering speed and \textbf{10$\times$} training speed.
\end{itemize}

\section{Related Works}
We review the three main areas of previous work: snapshot compressive imaging, NeRF, and efficient radiance field rendering techniques, which are the components most related to our work.

\noindent\textbf{Snapshot Compressive Imaging.} Early SCI image decoding and reconstruction methods primarily rely on regularized optimization approaches \cite{yang2020shearlet, liao2014generalized, yuan2015generalized, liu2018rank}. These methods iteratively estimate the compressed images captured from the SCI encoding systems by solving optimization problems. Various regularizers and priors, such as sparsity \cite{yang2020shearlet} and total variation (TV) \cite{yuan2015generalized}, have been applied in this context. Instead of traditional gradient descent, the majority of these approaches utilize the alternating direction methods of multipliers (ADMM) \cite{boyd2011distributed}, which provides favorable results and greater adaptability across different systems. Notable optimization-based techniques include DeSCI \cite{liu2018rank} and GAP-TV \cite{yuan2015generalized}. However, these methods are often hindered by long processing times and limited scalability for high-resolution images.

With the advancement of deep learning, recent SCI decoding and reconstruction methods have increasingly leveraged neural networks. Different network architectures, such as U-Net \cite{ronneberger2015u} and generative adversarial networks (GAN) \cite{NIPS2014_5ca3e9b1}, have been incorporated into SCI decoders. These deep learning models are typically trained on large-scale synthetic SCI datasets, including thousands of synthetic SCI measurements and masks, due to the difficulty of obtaining extensive real SCI data. The training procedure of these networks involves various loss functions, including conventional mean squared error (MSE), feature loss \cite{johnson2016perceptual}, and GAN loss \cite{miao2019net}. For example, Qiao et al. \cite{qiao2020deep} developed an end-to-end CNN (E2E-CNN) for compressed image retrieval, while Cheng et al. \cite{cheng2020birnat} employed bidirectional recurrent neural networks (BIRNAT) with a bi-directional residual network (RNN) \cite{he2016deep} for sequential video frame reconstruction. To address memory and time complexities in large-scale data and novel masks. RevSCI \cite{cheng2021memory} reduced the time and memory complexity during large-scale video SCI training by designing a multi-group reversible 3D CNN architecture. Additionally, plug-and-play fast and flexible denoising CNN (PnP-FFDNet \cite{yuan2020plug}) combined deep denoising networks with ADMM for faster, more flexible reconstruction, and fast video denoising network (FastDVDNet) \cite{yuan2021plug} further improved performance by introducing advanced denoising techniques. Wang et al. introduced spatial-temporal Transformers (STFormer) \cite{wang2022spatial} and EfficientSCI \cite{wang2023efficientsci, cao2024hybrid} to exploit spatial and temporal correlations using Transformer \cite{vaswani2017attention} architecture. Although these deep learning-based methods can achieve high-quality reconstructions, they often lack generalization ability when applied to real-world datasets since they rely on pre-trained models from synthetic data. Furthermore, these existing approaches are typically limited to decoding images corresponding to specific 2D masks, which poses challenges for reconstructing high frame-rate images or videos. In contrast, our methods employ test-time optimization, avoiding the generalization issues associated with pre-trained models, and they are capable of restoring high frame-rate multi-view consistent images from estimated 3D scenes.

\textbf{NeRF and Its Variants.} Introduced by Mildenhall et al. \cite{mildenhall2021nerf}, NeRF is a groundbreaking method for 3D scene representation, providing superior performance over traditional 3D scene representation techniques by accurately reconstructing novel views of the scene. Several variants have been proposed to adapt NeRF for more complex real-world applications. Some focus on scaling NeRF for large-scale scene reconstruction \cite{gu2021stylenerf, xiangli2022bungeenerf, turki2022mega, tancik2020fourier}, while others extend its capabilities for non-rigid object \cite{athar2022rignerf, gafni2021dynamic, peng2021animatable, peng2021neural}reconstruction. Additionally, NeRF-based approaches have been developed for high dynamic range (HDR) imaging \cite{Huang_2022_CVPR}, scene editing \cite{yang2021learning, yuan2022nerf, weder2023removing, wei2023clutter}, and scene appearance decomposition \cite{bi2020neural, boss2021nerd}. 

This section reviews NeRF-based methods relevant to our work. The original NeRF model requires precise camera poses as input, which is often unavailable in many image datasets. Most of the NeRF-based methods, including the original NeRF implementation, estimate camera poses from input images using SfM tools like COLMAP \cite{schonberger2016structure}. However, SfM may derive inaccurate camera poses or, even worse, fail on some of the input images. To overcome this limitation, various approaches optimize NeRF without relying on camera poses. For instance, Wang et al. propose NeRF-- \cite{wang2021nerf}, which jointly estimates the 3D scene and camera parameters (including intrinsic parameters and camera poses) during each training iteration. Jeong \cite{jeong2021self} introduces a self-calibration technique for camera poses, while iMAP \cite{sucar2021imap}, developed by Sucar et al., integrates NeRF with simultaneous localization and mapping (SLAM) to optimize both scene and camera trajectory. Meng et al. \cite{meng2021gnerf} apply GANs to enhance NeRF optimization without accurate poses. Other research focuses on refining inaccurate camera poses during optimization. The bundle-adjusted radiance fields (BARF) \cite{lin2021barf} method proposed by Lin et al. uses a coarse-to-fine registration strategy and dynamic low-pass filtering to optimize scene representation alongside camera poses. Building on BARF, Wang et al. propose BAD-NeRF \cite{Wang_2023_CVPR}, which optimizes both camera trajectories and the NeRF network by interpolating camera poses to simulate motion blur and estimate clearer scenes from blurred images. 

\textbf{Efficient Radiance Field Rendering.} NeRF employs a multi-layer perceptron (MLP) combined with volumetric rendering to implicitly represent scene radiance fields, encompassing both geometry and appearance. While NeRF has demonstrated significant success in 3D scene estimation and novel-view synthesis, it suffers from slow training and rendering speed. To overcome these limitations, several approaches have been proposed, such as grid-based methods like TensoRF \cite{chen2022tensorf}, Plenoxels \cite{fridovich2022plenoxels}, and HexPlane \cite{cao2023hexplane}, as well as hash-based solutions like InstantNGP \cite{mueller2022instant}. However, achieving real-time rendering without compromising image quality remains a significant challenge. The 3DGS framework, introduced by Kerbl et al. \cite{kerbl20233d}, strikes a balance between speed and quality by utilizing a purely explicit, point-based scene representation. In contrast to NeRF, 3DGS represents scenes with a group of 3D Gaussians and employs a tile-based rasterization scheme for Gaussian splats, enabling real-time high-quality rendering for high-resolution novel-view images. However, 3DGS heavily relies on accurate camera poses and sparse point clouds, typically obtained from COLMAP, to initialize Gaussians at the start of training. 

Subsequent works have introduced various refinements to address this limitation. InstaSplat \cite{fan2024instantsplat} utilizes DUSt3R \cite{wang2024dust3r}, a Transformer-based dense stereo model, to generate dense point clouds from sparse-view images. However, its focus on stereo matching leads to less accurate camera parameters. In contrast, VGGSfM \cite{wang2024vggsfm} incorporates a fully differentiable SfM pipeline with deep learning integration at each stage, providing highly accurate camera poses that benefit downstream tasks like 3D reconstruction \cite{HH}. Other studies try to improve the 3DGS densification strategy \cite{bulo2024revising, zhang2024pixel, ye2024absgs}. The original 3DGS applies a heuristic densification strategy for growing Gaussians, which has been identified as suboptimal. To enhance this process, these studies refine error criteria to densify Gaussians more effectively. Notably, recent approaches have reformulated Gaussian updates as state transformation within MCMC samples \cite{kheradmand20243d}, stabilizing training by recalculating opacity and scale values when duplicating Gaussians.
\section{Mathematical Model of Video SCI}


The formation process of a video SCI system is similar to that of a blurry image. The difference is that the captured images $\mathcal{X}=\{\mathbf{X}_i\in\mathbb{R}^{H\times W}\}^{N_I}_{i=1}$ of a video SCI system are modulated by $N_I$ binary masks $\mathcal{M}=\{\mathbf{M}_i\in\mathbb{R}^{H\times W}\}_{i=1}^{N_I}$ throughout the exposure time, where both $H$ and $W$ are image height and width, respectively. For practical hardware implementation, those masks are achieved by displaying different 2D patterns on the Digital Micro-mirror Device (DMD) or a spatial light modulator, e.g., liquid crystal. The image sensor then accumulates the modulated photons across exposure time to a compressed/coded image. The number of masks or different patterns on the DMD within the exposure time determines the number of coded frames, i.e. the temporal compression ratio (CR). Due to mask modulation, the $N_I$ virtual images can be recovered from a single compressed image alone by solving an ill-posed inverse problem.
%
The whole imaging process can be described formally as follows:
\begin{equation}\label{eq_formation}
	\mathbf{Y}=\sum_{i=1}^{N_I}\mathbf{X_i}\odot\mathbf{M_i}+\mathbf{Z},
\end{equation}
where $\mathbf{Y}, \mathbf{X}_i \in \mathbb{R}^{H \times W}$ are the compressed captured image and the $i^{th}$ virtual image within exposure time, respectively, $N_I$ is the temporal CR, $\odot$ denotes element-wise multiplication, and $\mathbf{Z} \in \mathbb{R}^{H \times W}$ is the measurement noise. The individual pixel value in the binary mask is generated randomly. For masks $N_I$ throughout the exposure time, the probability of assigning 1 to the same pixel location is fixed and defined as overlapping ratio. The optimal overlapping ratio is selected through an ablation study. 

\section{Method}

\begin{figure*}[!htbp]
  \centering
    \includegraphics[width=\linewidth]{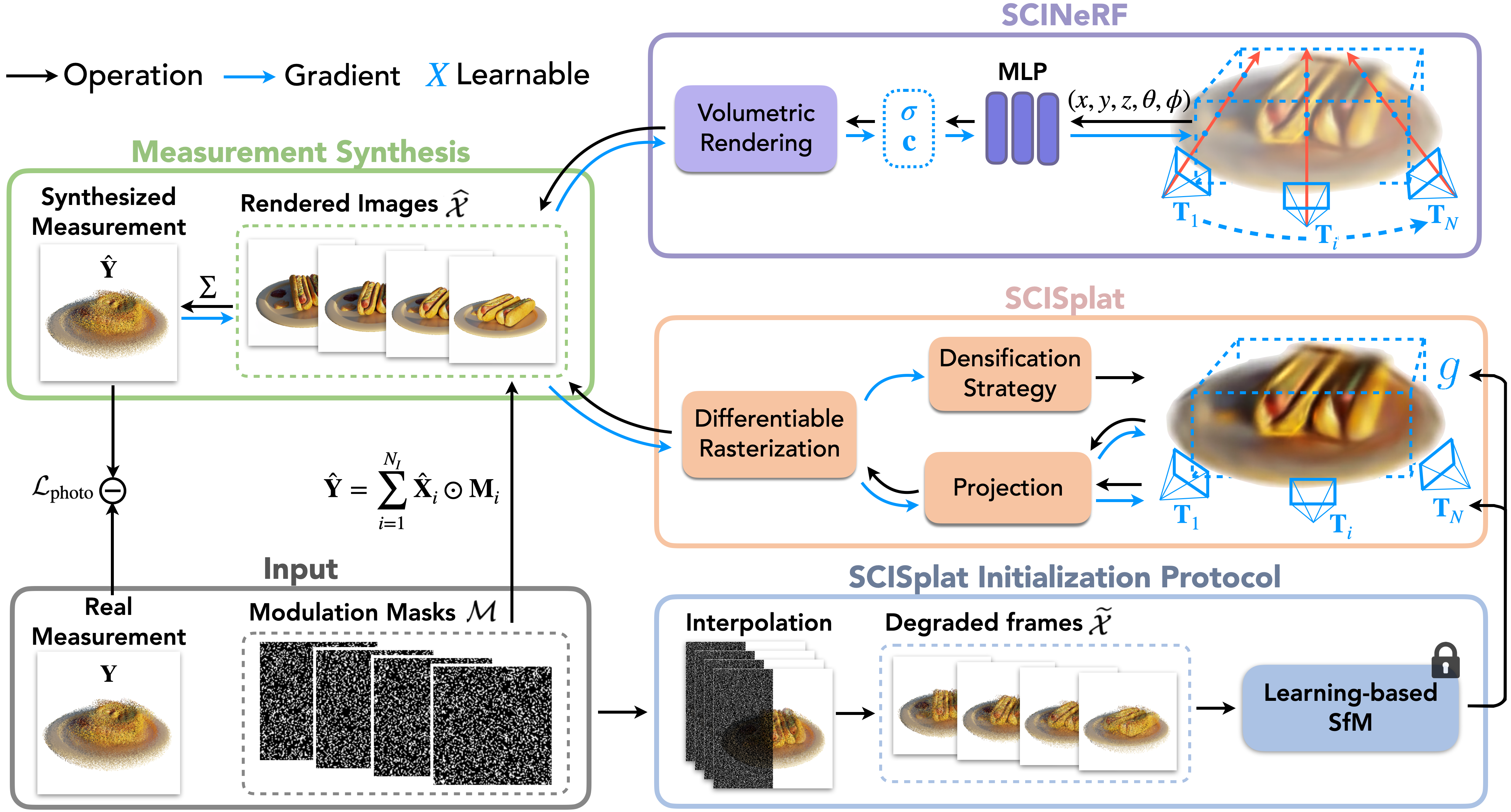}
    \caption{\textbf{Overview of Proposed Methods.} Both methods take the real SCI measurement $\mathbf{Y}$ and modulation masks $\mathcal{M}$ as input to recover the compressed images and the underlying 3D scene structure.
For \textbf{SCINeRF}, camera poses $\mathbf{T}_i$ are constrained by a spline. The scene information, including scene volumetric density $\sigma$ and RGB color $\mathbf{c}$, is encoded in a lightweight MLP, which is then used to render the compressed multi-view images $\hat{\mathcal{X}}$ through volumetric rendering.
To improve the 3D representation quality and training/rendering speed, we further propose \textbf{SCISplat} by incorporating 3DGS with SCI. For SCISplat, a set of degraded frames $\widetilde{\mathcal{X}}$ are first reconstructed from the real measurement $\mathbf{Y}$ and modulation masks $\mathcal{M}$ using pixel interpolation. These frames are then fed into a learning-based Structure-from-Motion (SfM) module to generate an initial coarse point cloud $\mathcal{Q}$ and estimate rough camera poses $\mathbf{T}_i$. These outputs serve as the initialization for the 3D Gaussian $g$ parameters. Subsequently, the compressed images $\hat{\mathcal{X}}$ are rendered via differentiable rasterization.
In both methods, the scene representation and camera poses are jointly optimized primarily by minimizing the photometric loss between the synthesized measurement $\hat{\mathbf{Y}}$ (from the rendered multi-view images from NeRF and 3DGS) and the real SCI measurement $\mathbf{Y}$.}
    \label{proposed_method}
    
\end{figure*}

Our proposed method takes a single compressed image and encoding masks as input and recovers the underlying 3D scene structure, as well as camera poses. Then, we can render high frame-rate images from the reconstructed 3D scene. To achieve this, we first exploit NeRF and then 3DGS as the underlying 3D scene representation. We follow the image formation model of video SCI to synthesize snapshot compressive images from NeRF and 3DGS. By maximizing the photometric consistency between the synthesized image and the actual SCI measurement, we optimize both 3D scene representation and camera poses. 
%
%
An overview of our method is shown in Fig.~\ref{proposed_method}, whose components are detailed as follows.

\subsection{Preliminaries}
\subsubsection{NeRF}

Given a set of multi-view images and their corresponding camera intrinsic and extrinsic parameters, NeRF maps the input image pixels to rays in 3D space. For each ray, points are sampled along its path, and the 5D features of these points, which consist of their 3D position and the 2D viewing direction, are used as input. A multilayer perceptron (MLP) then predicts the volumetric density \( \sigma \) and the view-dependent RGB color \( \mathbf{c} \) for each sampled point. By incorporating the viewing direction into the color prediction, NeRF effectively handles phenomena such as specular reflections and complex lighting interactions within the scene.

Once the density and color of each sampled point are predicted, NeRF uses differentiable volumetric rendering to integrate these values along the ray, ultimately synthesizing the pixel intensity \( C \) for the rendered image. This integration process can be formally expressed as follows:
\begin{equation}
    C(\mathbf{r}) = \int_{t_n}^{t_f} T(t) \sigma(\mathbf{r}(t)) \mathbf{c}(\mathbf{r}(t), \mathbf{d}) \, dt,
\end{equation}
where \( t_n \) and \( t_f \) are the near and far bounds for volumetric rendering, respectively. The function \( \mathbf{r}(t) \) represents the 3D point sampled along the ray \( \mathbf{r} \) at the distance \( t \) from the center of the camera. The term \( \sigma(\mathbf{r}(t)) \) denotes the predicted density of the sampled point \( \mathbf{r}(t) \) from the MLP. The accumulated transmittance \( T(t) \), which accounts for the opacity along the ray from \( t_n \) to \( t \), is defined as: $ T(t) = \exp\left(-\int_{t_n}^{t} \sigma(\mathbf{r}(s)) \, ds\right)$, where \( \mathbf{d} \) is the viewing direction in the world coordinate frame, and \( \mathbf{c}(\mathbf{r}(t), \mathbf{d}) \) is the predicted color of the sampled point \( \mathbf{r}(t) \).

To train the NeRF MLP network, the photometric loss is employed, which is the mean squared error (MSE) between the rendered pixel intensity and the real captured intensity:
\begin{equation}
  \mathcal{L} = \sum_{\mathbf{r} \in \mathcal{R}} \left\| \hat{C}(\mathbf{r}) - C(\mathbf{r}) \right\|_2^2,
\end{equation}
where \( \hat{C}(\mathbf{r}) \) and \( C(\mathbf{r}) \) denote the rendered and real captured pixel intensities for the ray \( \mathbf{r} \), respectively, and \( \mathcal{R} \) is the set of sampled rays.

\subsubsection{3DGS}
One of the major drawbacks of NeRF is that its implicit scene representation and volumetric rendering process incur high computational costs and long training/rendering times, making it unsuitable for fast training and real-time rendering. To overcome this, 3DGS leverages 3D Gaussian as a more efficient scene representation. A set of 3D Gaussians $g=\{\mathbf{g}_i\}_{i=1}^{M}$, parameterized by their mean position $\boldsymbol{\mu}_i \in \mathbb{R}^3$, 3D covariance $\mathbf{\Sigma}_i \in \mathbb{R}^{3 \times 3}$, opacity $o_i \in \mathbb{R} $, and color $\mathbf{c}_i \in \mathbb{R}^3$, is introduced to faithfully represent the 3D scene. 
The distribution of each Gaussian $\mathbf{g}_i$ is defined as:
\begin{equation}
    \mathbf{g}_i(\mathbf{x}) = \exp\left({-\frac{1}{2}(\mathbf{x}-\boldsymbol{\mu}_i)^{\top}\mathbf{\Sigma}_i^{-1}(\mathbf{x}-\boldsymbol{\mu}_i)}\right),
    \label{gaussian_definition}
\end{equation}

To ensure that the 3D covariance $\mathbf{\Sigma}_i$ remains positive semi-definite, which is physically meaningful, and to reduce optimization difficulty, 3DGS represents covariance matrix using a scale matrix $\mathbf{S}_i \in \mathbb{R}^{3\times3}$ (diagonal matrix) and rotation matrix $\mathbf{R}_i \in \mathbb{R}^{3\times3}$:
 \begin{equation}
\mathbf{\Sigma}_i = \mathbf{R}_i \mathbf{S}_i \mathbf{S}_i^\top \mathbf{R}_i^\top.
\end{equation}

Similar to NeRF, the input to 3DGS also consists of a series of multi-view images \( \mathcal{X} = \{\mathbf{X}_i \in \mathbb{R}^{H \times W}\}_{i=1}^N \) of the target 3D scene, along with their corresponding intrinsic and extrinsic camera parameters. These parameters are represented by camera projection matrices \( \mathcal{P} = \{\mathbf{P}_i \in \mathbb{R}^{3 \times 4}\}_{i=1}^N \). Additionally, at the beginning of 3DGS training, a sparse point cloud \( \mathcal{Q} = \{\mathbf{Q}_i \in \mathbb{R}^3\}_{i=1}^M \) of the target 3D scene, typically obtained through Structure-from-Motion (SfM), is used to initialize the Gaussians \( g \).

To render multi-view images, 3DGS employs a differentiable Gaussian rasterization process. In this process, 3D Gaussians are projected onto the 2D image plane based on a given camera pose \( \mathbf{T}_i = \{\mathbf{R}_{c,i}, \mathbf{t}_{c,i}\} \), where \( \mathbf{R}_{c,i} \in \mathbb{R}^{3 \times 3} \) is the camera rotation matrix and \( \mathbf{t}_{c,i} \in \mathbb{R}^3 \) is the translation vector. The projection is described by the following equation~\cite{kerbl20233d}:  
\[
\mathbf{\Sigma}'_i = \mathbf{J}_i \mathbf{R}_{c,i} \mathbf{\Sigma}_i \mathbf{R}_{c,i}^\top \mathbf{J}_i^\top,
\]  
where \( \mathbf{\Sigma}'_i \in \mathbb{R}^{2 \times 2} \) is the 2D covariance matrix, \( \mathbf{J}_i \in \mathbb{R}^{2 \times 3} \) is the Jacobian matrix of the affine approximation of the projective transformation.

Then, the image pixels are rendered by rasterizing these sorted 2D Gaussians based on their depths: 
\begin{equation}
C = \textstyle \sum_{i=1}^{M} \mathbf{c}_i \alpha_i \prod_{j=1}^{i-1} (1 - \alpha_j),
\end{equation}
where $C$ represents the pixel in the rendered multi-view images  $\widehat{\mathcal{X}}=\{\widehat{\mathbf{X}}_i\in\mathbb{R}^{H\times W}\}_{i=1}^{N}$, and $\alpha_i$ denotes the alpha value computed by evaluating the 2D covariance $\mathbf{\Sigma}^{'}_{i}$ multiplied with the learnable Gaussian opacity $o_i$:
 \begin{equation}
\alpha_i = o_i \exp(-\sigma_i), \quad \sigma_i = \frac{1}{2} \Delta_i^\top \mathbf{\Sigma}_i^{'-1} \Delta_i,
\end{equation}
where $\Delta_i\in\mathbb{R}^2$ is the offset between the pixel center and the projected 2D Gaussian center.

The 3D Gaussians $g$ are updated by minimizing the photometric loss computed between rendered images $\mathcal{\widehat{X}}$ and real captured images $\mathcal{X}$:
\begin{equation}
\begin{aligned}
      g^* &= \argmin_{g} \mathcal{L}_{\text{photo}}(\widehat{\mathcal{X}}, \mathcal{X}), \\
      \mathcal{L}_{\text{photo}} &= (1-\lambda_\text{D-SSIM})\cdot\mathcal{L}_1 + \lambda_{\text{D-SSIM}} \cdot \mathcal{L}_{\text{D-SSIM}}.
\end{aligned}
\label{photo_loss}
\end{equation}
where $\mathcal{L}_1$ is the average L1 distance, and $\mathcal{L}_{\text{D-SSIM}}$ is the Structural Similarity Index Metric (SSIM)~\cite{wang2004image}. As in Zhao et al.~\cite{image_loss}, $\lambda_{\text{D-SSIM}}$ is set to 0.2.




\subsection{SCINeRF: NeRF-based SCI decoding}

Given the image formation model of the SCI video SCI, we first apply NeRF as the underlying 3D scene representation. 
%
%
According to Eq. \ref{eq_formation}, we can interpret the SCI measurement $\mathbf{Y}$ as compressed multi-view images. However, directly reconstructing scenes from this single image is challenging due to the absence of camera poses. 
Given that SCI measurements are basically compressed frames in the form of a single 2D image, traditional SfM techniques, which are commonly employed in NeRF-based methods, are not directly applicable for estimating camera poses for individual frames. 

To address the issue of unknown poses, we adopt a strategy inspired by previous work \cite{lin2021barf,Wang_2023_CVPR}, in which the training procedure starts from the initialized inaccurate poses and simultaneously optimize NeRF and poses during training. Considering that the compressed multi-view images are captured within a short exposure time, we assume a linear camera trajectory during the imaging process. Consequently, we can obtain virtual camera poses using linear interpolation, which can be expressed as:
%
\begin{equation}\label{linear_pose_assumption}
    \mathbf{T}_i = \mathbf{T}_1\exp({\frac{i}{N}\log(\mathbf{T}_1^{-1}\mathbf{T}_N})),
\end{equation}
where $\mathbf{T}_i$ is the camera pose of $i^{th}$ frame, $N$ denotes the number of compressed frames, $\mathbf{T}_1\in \mathbf{SE}(3)$ and $\mathbf{T}_N\in \mathbf{SE}(3)$ are the poses of the first and last frames, respectively. They represent the starting and ending pose of the camera trajectory. Both $\mathbf{T}_1$ and $\mathbf{T}_N$ are initialized as quasi-identity matrices with small random perturbations to the translation components. The camera intrinsic parameters, including the camera focal length and the principle point of camera, come from the camera intrinsic calibration.

\subsection{SCISplat: 3DGS-based SCI decoding}

To improve SCINeRF's training and rendering speed and scene reconstruction quality, we further explore the potential of utilizing 3DGS as the 3D representation method for SCI images. SCISplat aims to reconstruct the target 3D scene and render high-quality encoded multi-view images in real-time. To properly train SCISplat, we follow the SCINeRF joint optimization strategy. 
We first propose a novel initialization protocol to estimate coarse point clouds and camera poses from a single SCI measurement to kick off the training procedure. Then, we jointly optimize Gaussians and camera poses when training SCISplat. 
Furthermore, to stabilize the training process, we leverage the MCMC \cite{kheradmand20243d} strategy for the densification of Gaussians. 

\subsubsection{Initialization Protocol} 
To initialize the training of 3DGS, we need not only the camera poses but also an initial point cloud that serves as a coarse approximation of the target 3D scene. Most existing 3DGS-based methods rely on COLMAP \cite{schonberger2016structure} to estimate 3D point clouds given the input multi-view images via SfM, while it brings extensive challenges for the SCI images since COLMAP will fail in such cases. To overcome these challenges, we propose a novel initialization protocol for our 3DGS-based approach.

Inspired by Wang et al. \cite{wang2023efficientsci}, we first normalize the real SCI measurement $\mathbf{Y}$ by the sum of all modulation masks $\mathbf{M_i}$:
\begin{equation}
     \overline{\mathbf{Y}}= \textstyle\mathbf{Y} \oslash \sum_{i=1}^{N}\mathbf{M}_i,
     \label{degraded_frame1}
\end{equation}
where $\overline{\mathbf{Y}}$ is the normalized measurement, and $\oslash$ denotes element-wise division. Then we can obtain the degraded frames $\mathcal{\widetilde{X}}=\{\mathbf{\widetilde{X}}_i\in\mathbb{R}^{H\times W}\}_{i=1}^{N_I}$ by interpolating the normalized measurement after modulated by a filtered version of each mask $\mathbf{M}_i\odot\mathbf{B}_i$:

\begin{equation}\label{degraded_frame2}
\begin{split}
    \mathbf{\widetilde{X}}_i &= \text{Interp}(\overline{\mathbf{Y}} \odot (\mathbf{M}_i \odot\mathbf{B}_i)),  \\
    \mathbf{B}_i(x,y) &= 
    \begin{cases} 
    1, & \text{if \ } \mathbf{M}_i(x,y) >= \tau \\
    0, & \text{otherwise}
    \end{cases},
\end{split}
\end{equation}
where $\mathbf{B}_i$ is a selection matrix that only preserves the value of $\mathbf{M}_i$ positioned at pixel coordinate $(x,y)$ when its value exceeds the threshold $\tau$. For synthetic datasets, the 2D masks are binary, which means the mask value on each pixel location has either a value of 0 or 1. In this case, we set $\tau=1$. For the real data, due to the existence of measurement noise, the mask values are not strictly binary, at which we carefully select the value of $\tau$ to be 0.8. The effects of different selected $\tau$ values are investigated in the ablation study.

As visualized in Fig.~\ref{proposed_method}, large parts of these degraded frames $\mathcal{\widetilde{X}}$ are contaminated by noise, which makes traditional feature tracking approaches fail easily. Therefore, classical SfM methods like COLMAP cannot be applied directly here, as most previous 3DGS-based frameworks. However, with the recent development of learning-based SfM methods, obtaining a decent guess from these noise images is possible. Specifically, we use VGGSfM~\cite{wang2024vggsfm} denoted as $f_\theta(\cdot)$, a fully differentiable SfM pipeline, to directly get camera projection matrix estimates $\mathcal{P}=\{\mathbf{P}_i \in \mathbb{R}^{3\times4}\}^{N_I}_{i=1}$ and an initial point cloud $\mathcal{Q}=\{\mathbf{Q}_i\in\mathbb{R}^{3}\}_{i=1}^{N_Q}$ from these degraded frames $\mathcal{\widetilde{X}}$, whose deep point tracker is relatively robust to noise presented:
\begin{equation}
    \mathcal{P, Q}=f_\theta(\mathcal{\widetilde{X}}).
\end{equation}
It is noteworthy that each projection matrix $\mathbf{P}_i$ consists of extrinsic $\mathbf{T}_i \in \mathbf{SE}(3)$ representing the camera poses, and intrinsic $\mathbf{K}\in\mathbb{R}^{3\times3}$. $\mathbf{T}_i$ will be optimized, but $\mathbf{K}$ will remain fixed for all frames throughout training.

Given that the initial point cloud $\mathcal{Q}$ derived from degraded frames $\mathcal{\widetilde{X}}$, it will inevitably contain noisy points that will impair the final rendering quality. In addition, having many points at the start of training will quickly introduce artifacts due to inaccurate camera poses $\mathbf{T}_i$. Thus, we uniformly downsample the initial point cloud $\mathcal{Q}$ to a certain number $n$ as $\mathcal{Q}_n$ to mitigate the formation of noise artifacts at the early stage. However, having too few points also challenges the ability of the densification strategy to faithfully reconstruct the scene, given that we are recovering the scene from a compressed measurement. Therefore, choosing a suitable number of initial points is vital to the final quality of the reconstruction, which is further investigated in the ablation study. After downsampling, the initial set of $n$ Gaussians $g=\{\mathbf{g_i}\}_{i=1}^{n}$ are placed at the locations of these downsampled points $\mathcal{Q}_n$.

\subsubsection{Optimization Strategy}
When training 3DGS, Gaussians are progressively adjusted to express fine details. 3DGS automatically introduces new Gaussians to regions with larger reconstruction loss, where the initial Gaussians fail to reconstruct these areas. In our proposed framework, instead of conventional adaptive density control (ADC) introduced in the original 3DGS \cite{kerbl20233d}, we employ the MCMC strategy \cite{kheradmand20243d} for the reasons below. In the original ADC strategy, the composed opacity values are larger after cloning or splitting Gaussians, making the scene appear slightly brighter. This inconsistent update introduces spiky appearance changes in the scene, which sometimes can make camera poses drift to sub-optimal locations due to the unstable gradient flow from photometric loss, thus collapsing the whole reconstruction. Moreover, due to pixel ambiguity in fitting an SCI measurement, suddenly having high opacity for Gaussians will quickly introduce noise in the reconstructed viewpoint, as indicated by Eq.~\ref{eq_formation}. With this ill-posed image formation model, a sub-optimal solution could quickly occur where one pixel has a high pixel value close to the sum while others appear much darker. In this case, that pixel will appear as noise on the reconstructed images. Recalling the original strategy, suddenly making individual Gaussian opacity high will undoubtedly encourage the reach of this local optimal and thus lead to noisy reconstruction. In contrast, the MCMC strategy corrects this opacity bias by recomputing the updated opacity values after densification, resulting in smoother training dynamics and effective noise suppression.

\subsection{Loss Function}

For both NeRF and 3DGS-based approaches, we face the same problem regarding the loss function that only one input SCI image is available per scene, while conventional NeRF and 3DGS compute loss on multi-view images directly. To achieve end-to-end training of our NeRF and 3DGS-based framework on a single SCI image, we follow the image formation model of video SCI indicated by Eq. \eqref{eq_formation} to transfer the multi-view images rendered from NeRF and 3DGS into a synthesized SCI measurement:
\begin{equation}
    \widehat{\mathbf{Y}} = \sum_{i=1}^{N_I}\widehat{\mathbf{X}}_i \odot \mathbf{M}_i,
\end{equation}
where $\widehat{\mathbf{Y}}\in\mathbb{R}^{H\times W}$ represents the synthesized SCI measurement. Here, we omit the measurement noise term $\mathbf{Z}$ in Eq. \ref{eq_formation} to facilitate the recovery of the originally captured image. 

Then for SCINeRF, we simply compute the MSE between the synthesized SCI measurement $\widehat{\mathbf{Y}}$ and the real captured measurement $\mathbf{Y}$ as the loss function:
\begin{equation}
    \mathcal{L}=\left\|\widehat{\mathbf{Y}}-\mathbf{Y}\right\|^{2}.
\end{equation}
Similarly, for 3DGS-based SCISplat, we compute the photometric loss between synthesized and real SCI measurement with two additive regularization terms:
\begin{equation}
\begin{split}
    \mathcal{L} = \mathcal{L}_{\text{photo}}(\widehat{\mathbf{Y}}, \mathbf{Y}) &+  \lambda_\text{o} \cdot \mathcal{L}_\text{o}(g) + \lambda_\text{s} \cdot \mathcal{L}_\text{s}(g), \\
    \mathcal{L}_\text{o}(g) = \Sigma_i|o_i|_1&, \mathcal{L}_\text{s}(g)= \Sigma_{ij}\left|\sqrt{eig_j(\Sigma_i)}\right|_1,
\end{split}
\end{equation}
where $\mathcal{L}_{\text{photo}}$ is the same photometric loss function as indicated by Eq.~\eqref{photo_loss}, and the latter two terms are scale term $\mathcal{L}_\text{s}$ and opacity term $\mathcal{L}_\text{o}$, in which $eig(.)$ denotes the j-th eigenvalue of covariance matrix. These two terms minimize the scale and opacity of the current Gaussians to encourage a lower number of effective Gaussians. With this loss, we can jointly optimize camera poses and 3D scenes represented by NeRF and 3DGS. This joint optimization strategy enables us to compensate for inaccuracies in the initial poses, thereby enhancing the reconstruction quality.

\section{experiments}
We validate our proposed method on synthetic and real datasets captured by our system, comparing its performance with state-of-the-art (SOTA) SCI image restoration approaches. Experimental results demonstrate that our method achieves superior restoration quality compared with existing techniques. We further prove that incorporating 3DGS with SCI delivers higher performance in image restoration quality and achieves significantly faster training and rendering speed.

\subsection{Experimental Setup}
\subsubsection{Synthetic Datasets}
To comprehensively evaluate the performance of our proposed method, we generate synthetic datasets derived from widely-used multi-view datasets, which are commonly employed in image-based rendering and novel-view synthesis works. We include data from scene \textit{Airplants} from LLFF dataset \cite{mildenhall2019local}, and \textit{Hotdog} from NeRF synthetic 360 dataset \cite{mildenhall2021nerf}, which are part of standard benchmarks in this domain. Additionally, to better simulate the capturing process along a trajectory of a real camera, we extend our dataset generation by utilizing the scene provided in the DeblurNeRF \cite{ma2022deblur} dataset using the Blender software. We synthesize new datasets based on four virtual scenes: \textit{Cozy2room, Tanabata, Factory}, and \textit{Vender}. For synthetic datasets, we uniformly apply a compression ratio of 8 to all data sets. To further test the adaptability of our proposed method to varying spatial resolutions, we introduce diverse resolution configurations across the datasets: For LLFF, the images are at a resolution of 512 $\times$ 512; for NeRF Synthetic 360, the resolution is set to 400 $\times$ 400; and for the Blender-synthesized scenes (i.e., \textit{Cozy2room, Tanabata, Factory, Vender}), a resolution of 600 $\times$ 400 is employed. This variation ensures that our proposed method's performance is rigorously validated across different resolution scales, reflecting the method's robustness and scalability in practical scenarios.

\begin{figure}[tbp]
  \centering
    \includegraphics[width=\linewidth]{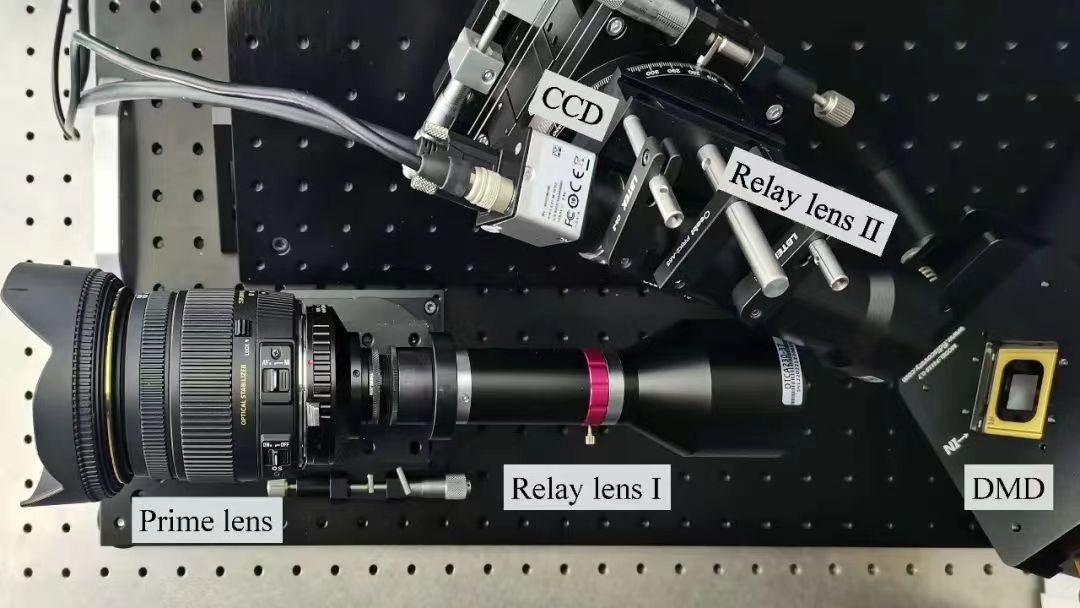}
    \caption{\textbf{Experimental setup for real dataset collection.} This SCI imaging system contains a CCD camera to record snapshot measurement, primary and rely lens, and a DMD to modulate input frames.}
    \label{experimental setup}
\end{figure}

\subsubsection{Real-world Datasets}
For the real-world dataset, we collect SCI measurements using two dedicated experimental setups. For grayscale data, we employed an SCI system that includes an iRAYPLE A5402MU90 camera in combination with a FLDISCOVERY F4110 DMD. Fig. \ref{experimental setup} shows this experimental setup, which illustrates the physical configuration and arrangement of the camera and DMD for data acquisition. For real color data, we used the HIKROBOT MV-CS020-10UC color camera with the same DMD. The SCI measurements are obtained at a resolution of 1024 $\times$ 768, with a compression ratio of 8. This real dataset serves as a critical evaluation to assess the practical viability of our method under realistic imaging conditions.

\subsubsection{Baseline Methods and Evaluation Metrics}
Since our method can render high-quality images from the estimated scene, we compared it with SOTA SCI image restoration methods to demonstrate its effectiveness. These baseline methods include model-based approaches, such as GAP-TV \cite{yuan2015generalized}, and deep learning-based frameworks such as PnP-FFDNet \cite{yuan2020plug}, PnP-FastDVDNet \cite{yuan2021plug}, and EfficientSCI \cite{wang2023efficientsci}. To ensure a fair comparison, we fine-tuned the EfficientSCI in advance using the same modulation masks applied in our datasets, optimizing its performance under the conditions of our experiments.

In addition to SCI image restoration comparisons, we evaluate the novel view synthesis capabilities of our method against baseline methods. For these comparisons, the restored images from prior SOTA SCI restoration methods are fed into the vanilla NeRF and original 3DGS models as input. However, we observed a critical limitation of existing restoration approaches. Due to their inability to recover high-quality details and the lack of multi-view consistency, it becomes infeasible to estimate accurate camera poses through COLMAP. To enable comparison, we provide NeRF and 3DGS with camera poses estimated from the ground truth images rather than from reconstructed images. Although this adjustment introduces a bias that is unfavorable to our method, it allows reasonable comparative results to be drawn. 

We use widely adopted metrics to ensure comprehensive analysis. These metrics include the structural similarity index measure (SSIM), the peak signal-to-noise ratio (PSNR) and the learned perceptual image patch similarity (LPIPS)~\cite{zhang2018unreasonable} metric. This diverse set of metrics ensures that both the perceptual and structural quality of the restored and synthesized images are adequately assessed. 

Additionally, since our method can optimize camera poses, we also evaluated its camera pose estimation capabilities by computing the absolute trajectory error, a commonly used evaluation metric in SLAM and visual odometry.

\subsubsection{Implementation Details}
Our method uses the PyTorch framework, providing a flexible and efficient environment for model development and optimizations. Specifically, for SCINeRF, we leverage the vanilla NeRF implementation from \cite{lin2020nerfpytorch} to model and represent the 3D scene. The optimization process employs two separate Adam optimizers \cite{kingma2017adam}, each dedicated to different components of the model: one for scene optimization and the other for camera pose optimization. For the scene optimization, the learning rate begins at $5 \times 10^{-4}$ and is decreased exponentially to $5 \times 10^{-5}$. The learning rate for pose optimization starts at $1 \times 10^{-3}$ and decays exponentially to $1 \times 10^{-5}$. We train the SCINeRF model for 100K-200K iterations using a batch size of 5000 rays.

For our SCISplat framework, we build on the implementation of the 3DGS pipeline \cite{kerbl20233d}, adapting it to our specific task. Similar to SCINeRF, we use Adam optimizer to optimize the Gaussian parameters and camera poses simultaneously. A key consideration in this process is the learning rate for the Gaussian parameters, which is scaled by the square root of 8, consistent with the square root rule \cite{andre2022sqr}, as we forward 8 images concurrently during each optimization step. For camera pose optimization, the learning rate begins at $5\times10^{-4}$ and exponentially decreases to $2.5\times10^{-7}$, allowing fine-tuning adjustments over the course of training. In addition, we set the maximum number of Gaussians used in the model at 100K, as defined by the MCMC \cite{kheradmand20243d} strategy, ensuring a robust and efficient exploration of the parameter space. All experiments are conducted on a single NVIDIA RTX 4090 GPU.

\subsection{Results}
\begin{table*}
        \caption{\textbf{Quantitative SCI image reconstruction comparisons on the synthetic dataset} The results are computed from the rendered images from estimated scenes via proposed SCINeRF and SCISplat, and recovered images from state-of-the-art SCI image restoration methods on six datasets: \textit{Airplants, Hotdog, Cozy2room, Tanabata, Factory} and \textit{Vender}. We apply PSNR, SSIM and LPIPS as evaluation metrics. The experimental results demonstrate that our SCINeRF and SCISplat can render images with higher quality than those from existing methods. The best results are shown in bold and the second-best results are underlined.}
	\setlength\tabcolsep{2pt}
	\parbox{\textwidth}{
		\resizebox{\linewidth}{!}{
		\begin{tabular}{c|ccc|ccc|ccc|ccc|ccc|ccc}
			
			\hline
			& \multicolumn{3}{c|}{Airplants} & \multicolumn{3}{c|}{Hotdog} & \multicolumn{3}{c|}{Cozy2room} & \multicolumn{3}{c|}{Tanabata} & \multicolumn{3}{c}{Factory}  & \multicolumn{3}{c}{Vender}\\
			& PSNR$\uparrow$ & SSIM$\uparrow$ & LPIPS$\downarrow$ & PSNR$\uparrow$ & SSIM$\uparrow$ & LPIPS$\downarrow$ & PSNR$\uparrow$ & SSIM$\uparrow$ & LPIPS$\downarrow$ & PSNR$\uparrow$ & SSIM$\uparrow$ & LPIPS$\downarrow$ & PSNR$\uparrow$ & SSIM$\uparrow$ & LPIPS$\downarrow$ & PSNR$\uparrow$ & SSIM$\uparrow$ & LPIPS$\downarrow$\\
			\hline
			
			GAP-TV &22.85 &0.406 &0.499 &22.35 &0.766 &0.318 &21.77 &0.432 &0.603 &20.42 &0.426 &0.625 &24.05 &0.566 &0.0.515 &20.00 &0.368 &0.688\\
			PnP-FFDNet &27.79 &0.912 &0.182 &29.00 &0.977 &0.051 &28.98 &0.892 &0.984 & 29.17 & 0.903 & 0.119 & 31.75 & 0.897 & 0.114 & 28.70 & 0.923 & 0.131\\
			PnP-FastDVDNet &28.18 & 0.909 & 0.175 & 29.93 & 0.972 & 0.052 & 30.19 & 0.913 & 0.079 & 29.73 & 0.933 & 0.098 & 32.53 & 0.916 & 0.105 & 33.17 & 0.940 & 0.045\\
			EfficientSCI &30.13 & {\underline{0.942}} & 0.112  & 30.75 & 0.956 & 0.046 & 31.47 & 0.932 & 0.047 & 32.30 & 0.958 & 0.060 & 32.87 & 0.925 & 0.070 & 33.17 & 0.940 & 0.045\\
            
			\hline
			
		      SCINeRF &{\underline{30.69}} &0.933 &{\underline{0.072}} &{\underline{31.35}} &{\underline{0.987}} &{\underline{0.031}} &{\underline{33.23}} &{\underline{0.949}} &{\underline{0.044}} &{\underline{33.61}} &{\underline{0.963}} &{\underline{0.037}} &{\underline{36.60}} &{\underline{0.963}} &{\underline{0.022}} & {\underline{36.40}} & {\underline{0.984}} & {\underline{0.029}}\\
                SCISplat &{\textbf{31.45}} &{\textbf{0.951}} &{\textbf{0.036}} &{\textbf{32.67}} &{\textbf{0.991}} &{\textbf{0.016}} &{\textbf{35.26}} & {\textbf{0.972}} &{\textbf{0.011}} &{\textbf{37.86}} &{\textbf{0.985}} &{\textbf{0.005}} &{\textbf{38.92}} &{\textbf{0.975}} &{\textbf{0.010}} & {\textbf{39.49}} & {\textbf{0.992}} & {\textbf{0.004}}\\
            \hline
			
	\end{tabular}}
    
	\label{table I}}

\end{table*}

\begin{table*}
        \caption{\textbf{Quantitative comparisons on novel-view synthesis} The results are the rendered novel-view images from SCINeRF and SCISplat and novel-view images from vanilla NeRF and 3DGS with reconstructed images from existing state-of-the-art methods. Since we cannot estimate accurate poses and point clouds from the SOTA outputs, we use ground truth images to estimate camera poses and point clouds of target scene instead. The experimental results show that our SCINeRF and SCISplat outperforms existing naive two-stage approaches.}
	\setlength\tabcolsep{2pt}
	\parbox{\textwidth}{
		\resizebox{\linewidth}{!}{
		\begin{tabular}{c|ccc|ccc|ccc|ccc|ccc|ccc}
			
			\hline
			& \multicolumn{3}{c|}{Airplants} & \multicolumn{3}{c|}{Hotdog} & \multicolumn{3}{c|}{Cozy2room} & \multicolumn{3}{c|}{Tanabata} & \multicolumn{3}{c|}{Factory} & \multicolumn{3}{c}{Vender}\\
			& PSNR$\uparrow$ & SSIM$\uparrow$ & LPIPS$\downarrow$ & PSNR$\uparrow$ & SSIM$\uparrow$ & LPIPS$\downarrow$ & PSNR$\uparrow$ & SSIM$\uparrow$ & LPIPS$\downarrow$ & PSNR$\uparrow$ & SSIM$\uparrow$ & LPIPS$\downarrow$ & PSNR$\uparrow$ & SSIM$\uparrow$ & LPIPS$\downarrow$ & PSNR$\uparrow$ & SSIM$\uparrow$ & LPIPS$\downarrow$\\
			\hline
			
			NeRF+GAP-TV &23.72 &0.468 &0.419 &23.80 &0.767 &0.284 &21.99 &0.502 &0.521 &20.91 &0.412 &0.553 &25.48 &0.685 &0.440 &21.68 &0.436 &0.553\\
			NeRF+PnP-FFDNet &26.72 &0.883 &0.224 &29.14 &0.972 &0.061 &30.15 &0.903 &0.091 & 28.00 & 0.902 & 0.140 & 31.96 & 0.885 & 0.182 &30.02 &0.934 &0.131\\
			NeRF+PnP-FastDVDNet &26.91 & 0.876 & 0.206 & 29.31 & 0.972 & 0.054 & 31.17 & 0.917 & 0.079 & 30.79 & 0.936 & 0.100 & 32.56 & 0.902 & 0.153 &31.30 &0.945 &0.115\\
			NeRF+EfficientSCI &28.62 & 0.914 & 0.159  & 29.82 & 0.976 & 0.052 & 31.79 & 0.935 & 0.064 & 31.35 & 0.935 & 0.064 &32.72 &0.894 &0.067 & 32.77 & 0.943 & 0.064 \\
            3DGS+GAP-TV &24.08 & 0.430 & 0.493  & 24.03 & 0.794 & 0.233 & 22.81 & 0.535 & 0.399 & 22.35 & 0.523 & 0.410 & 26.40 & 0.717 & 0.390 &22.20 &0.476 &0.436\\
            3DGS+PnP-FFDNet &28.51 & 0.917 & 0.194  & 30.22 & 0.979 & 0.074 & 31.01 & 0.915 & 0.092 & 32.67 & 0.947 & 0.100 & 31.76 & 0.925 & 0.112 &31.98 &0.954 &0.109\\
             3DGS+PnP-FastDVDNet &28.51 &0.917  & 0.194  & 30.62 & 0.980 & 0.069 & 31.48 & 0.916 & 0.093 & 33.47 & 0.953 & 0.092 & 32.09 & 0.935 & 0.095 &32.93 &0.958 &0.103\\
            3DGS+EfficientSCI &29.83 & {\underline{0.943}} & 0.099  & 30.42 & 0.908 & 0.062 & 31.93 & 0.949 & 0.055 & 33.73 & 0.970 & 0.064 & 34.40 & 0.939 & 0.097 &34.17 &0.982 &0.046\\
			\hline
			
			SCINeRF & {\underline{30.61}} & 0.938 & {\underline{0.076}} & {\underline{30.59}} & {\underline{0.981}} & {\underline{0.044}} & {\underline{33.26}} & {\underline{0.952}} & {\underline{0.035}} & {\underline{35.52}} & {\underline{0.981}} & {\underline{0.029}} & {\underline{36.55}} & {\underline{0.952}} & {\underline{0.016}} & {\underline{36.68}} & {\underline{0.984}} & {\underline{0.025}}\\
                SCISplat & {\textbf{30.64}} & {\textbf{0.945}} & {\textbf{0.038}} & {\textbf{32.79}} & {\textbf{0.991}} & {\textbf{0.036}} & {\textbf{34.53}} & {\textbf{0.970}} & {\textbf{0.013}} & {\textbf{37.02}} & {\textbf{0.986}} & {\textbf{0.011}} & {\textbf{38.54}} & {\textbf{0.970}} & {\textbf{0.010}} & {\textbf{38.96}} & {\textbf{0.992}} & {\textbf{0.005}}\\
            \hline
			
	\end{tabular}}

	\label{table II}}
\end{table*}
\begin{figure*}[!tbp]

\begin{minipage}[c]{1.0\textwidth}
\begin{minipage}[c]{\linewidth}
\centering
  \begin{minipage}[c]{0.135\linewidth}
  \centering
  \small
  \ \\ \ Measurement
  \end{minipage}
  ~
  \begin{minipage}[c]{0.135\linewidth}
  \centering
  \small
  \ \\ PnP-FFDNet
  \end{minipage}
  ~
  \begin{minipage}[c]{0.135\linewidth}
  \centering
  \small
  \ \\ \ EfficientSCI
  \end{minipage}
  ~
  \begin{minipage}[c]{0.135\linewidth}
  \centering
  \small
  \ \\ \ SCINeRF
  \end{minipage}
  ~
  \begin{minipage}[c]{0.135\linewidth}
  \centering
  \small
  \ \\ \ SCISplat
  \end{minipage}
  ~
  \begin{minipage}[c]{0.135\linewidth}
  \centering
  \small
  \ \\ \ Ground Truth
  \end{minipage}
\end{minipage}
\\
\begin{minipage}[c]{\linewidth}
\centering
  \begin{minipage}[c]{0.135\linewidth}
  \includegraphics[width=1\linewidth]{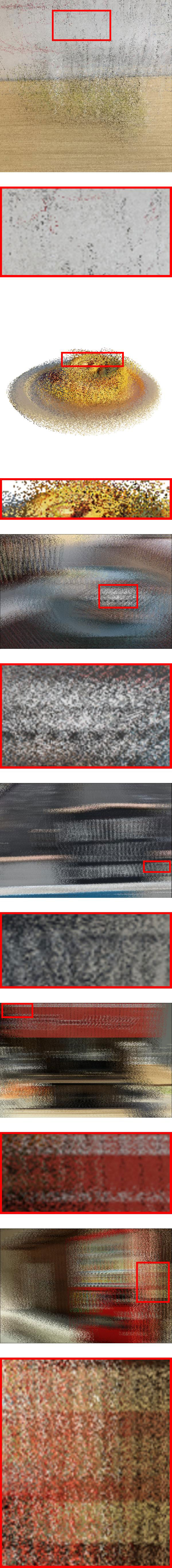}
  
  \end{minipage}
  ~
  \begin{minipage}[c]{0.135\linewidth}
  \includegraphics[width=1\linewidth]{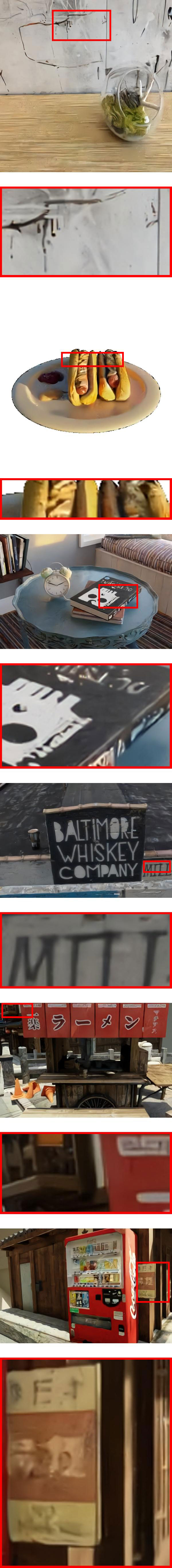}
  \end{minipage}
  ~
  \begin{minipage}[c]{0.135\linewidth}
  \includegraphics[width=1\linewidth]{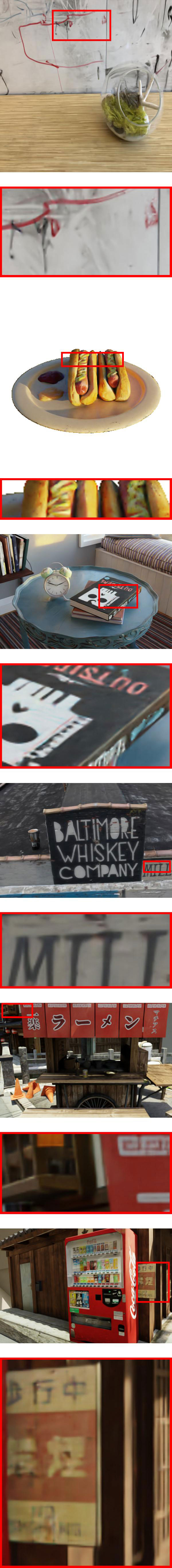}
  \end{minipage}
  ~
  \begin{minipage}[c]{0.135\linewidth}
  \includegraphics[width=1\linewidth]{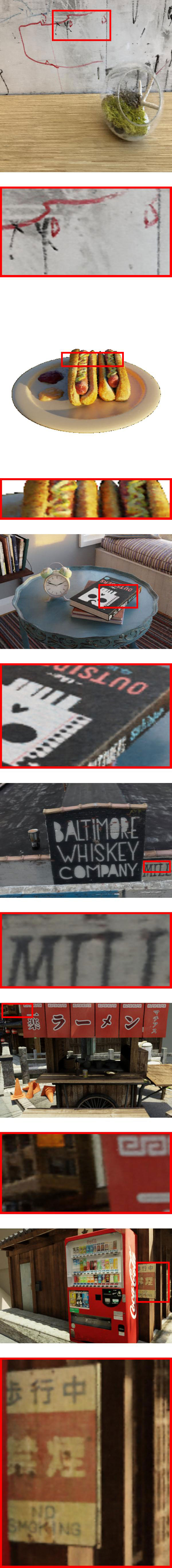}
  \end{minipage}
  ~
  \begin{minipage}[c]{0.135\linewidth}
  \includegraphics[width=1\linewidth]{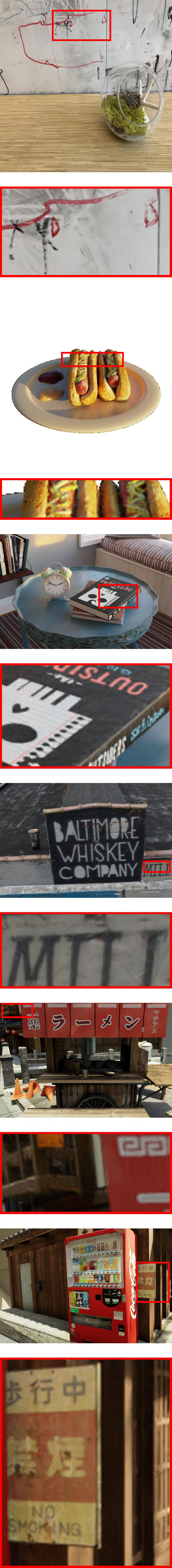}
  \end{minipage}
  ~
  \begin{minipage}[c]{0.135\linewidth}
  \includegraphics[width=1\linewidth]{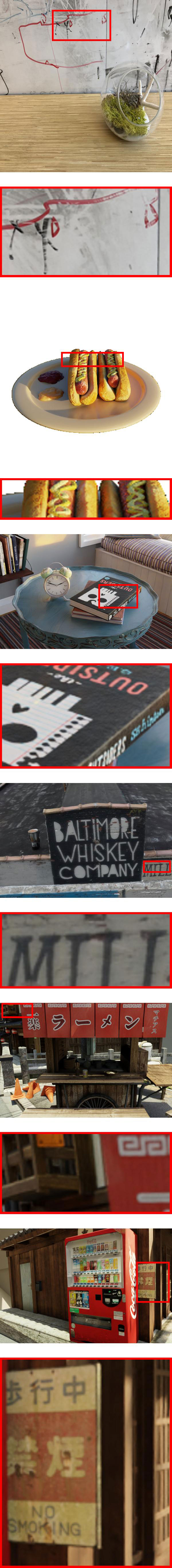}
  \end{minipage}
  
\end{minipage}
\end{minipage}
\\

   \caption{\textbf{Qualitative evaluations of our methods against SOTA SCI image restoration methods on the synthetic dataset.} Top to bottom shows the results for different scenes, including \textit{Airplants, Hotdog, Cozy2room, Factory, Tanabata,} and \textit{Vender}. The experimental results demonstrate that our SCINeRF and SCISplat achieve superior performance from a single SCI image (the far left column).
   }
   \label{fig4}
\end{figure*}
\subsubsection{Synthetic Dataset}
We compare the performance of our proposed method against SOTA techniques in the context of SCI image reconstruction. Quantitative results in synthetic datasets, summarized in Table \ref{table I}, provide strong empirical evidence for the efficacy of our SCINeRF framework in estimating and representing high-quality 3D scenes from a single SCI measurement. Specifically, SCINeRF achieves an average PSNR of 33.64dB, an average SSIM of 0.963, and a mean LPIPS lower than 0.02. These results show a significant improvement of over 2dB in PSNR compared to EfficientSCI, the best-performing method among the SOTA SCI image reconstruction approaches. 
%

Based on these results, we introduce our 3DGS-based SCISplat approach, which delivers even better performance. As shown in Table \ref{table I}, SCISplat achieves an average PSNR exceeding 35.94dB, an average SSIM above 0.977, and a mean LPIPS of 0.0136. This represents a gain of over 2.3dB in PSNR compared to SCINeRF and over 4.3dB compared to the best SOTA SCI image reconstruction method. 

Fig. \ref{fig4} illustrates the qualitative results of our methods compared to SOTA SCI image restoration approaches. The PnP-FFDNet approach can restore the overall structure of the target scene but lacks the ability to recover finer details. While EfficientSCI benefits from its powerful transformer-based self-attention mechanism and can reconstruct broadly realistic scenes, its results tend to exhibit "over-smoothing" leading to a loss or a distortion of critical details in reconstructed images. This limitation becomes particularly apparent in failing to recover fine textures, patterns, or characters, which we attribute to the generalization gap in pre-trained deep-learning models, even after being fine-tuned.

In contrast, our methods, SCINeRF and SCISplat excel in recovering intricate details by recovering textures and characters with high fidelity. SCISplat, in particular, demonstrates unparalleled reconstruction quality in scenes with complex visual features, providing clear and visually accurate outputs that significantly outperform existing SOTA methods. These findings underscore the robustness and versatility of our approaches in handling diverse and challenging imaging scenarios. 

To assess the performance of our proposed method in the novel-view synthesis task, we conducted experiments on novel-view synthesis. Unlike prior SOTA SCI image reconstruction methods that recover only 2D images, our approach incorporates 3D scene representation frameworks, making it inherently more suited for novel-view synthesis. Since the SOTA methods have no 3D representation capabilities, we design a naive two-stage pipeline for the baseline comparison where reconstructed 2D images from SOTA methods are fed into vanilla NeRF and 3DGS models to render novel-view images. However, we noticed that the COLMAP, which is applied by most NeRF-based and 3DGS-based approaches for camera pose estimation and point cloud extraction, fails to process these reconstructed images effectively due to their lack of multi-view consistency and insufficient scene details. Consequently, we use camera poses and point clouds derived from ground truth images as the input of NeRF and 3DGS. Although this setup introduces a favorable bias for the baseline methods, our approach still significantly outperforms these naive pipelines.

\begin{figure*}[!tbp]

\begin{minipage}[c]{1.0\textwidth}
\begin{minipage}[c]{\linewidth}
\centering
  \begin{minipage}[c]{0.14\linewidth}
  \centering
  \small
  \ \\ \ Measurement
  \end{minipage}
  ~
  \begin{minipage}[c]{0.14\linewidth}
  \centering
  \small
  \ \\ 3DGS+\\PnP-FFDNet
  \end{minipage}
  ~
  \begin{minipage}[c]{0.14\linewidth}
  \centering
  \small
  \ \\ \ 3DGS+\\EfficientSCI
  \end{minipage}
  ~
  \begin{minipage}[c]{0.14\linewidth}
  \centering
  \small
  \ \\ \ SCINeRF
  \end{minipage}
  ~
  \begin{minipage}[c]{0.14\linewidth}
  \centering
  \small
  \ \\ \ SCISplat
  \end{minipage}
  ~
  \begin{minipage}[c]{0.14\linewidth}
  \centering
  \small
  \ \\ \ Ground Truth
  \end{minipage}
\end{minipage}
\\
\begin{minipage}[c]{\linewidth}
\centering
  \begin{minipage}[c]{0.14\linewidth}
  \includegraphics[width=1\linewidth]{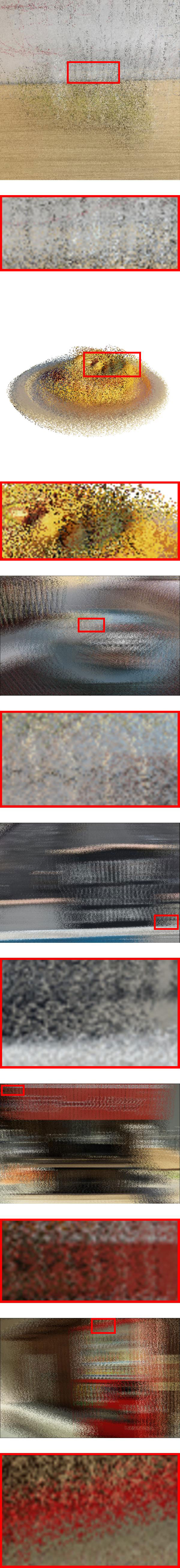}
  
  \end{minipage}
  ~
  \begin{minipage}[c]{0.14\linewidth}
  \includegraphics[width=1\linewidth]{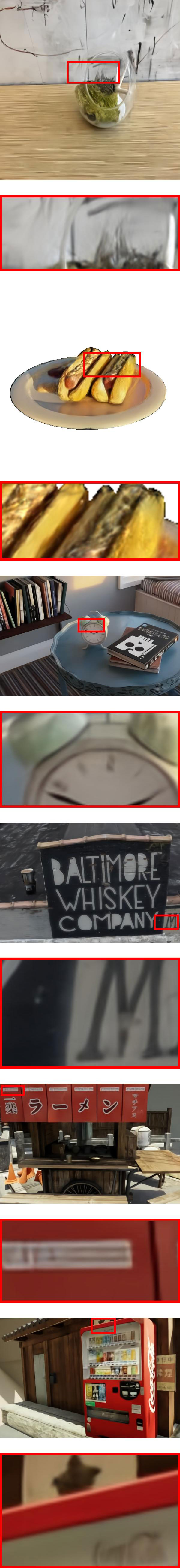}
  \end{minipage}
  ~
  \begin{minipage}[c]{0.14\linewidth}
  \includegraphics[width=1\linewidth]{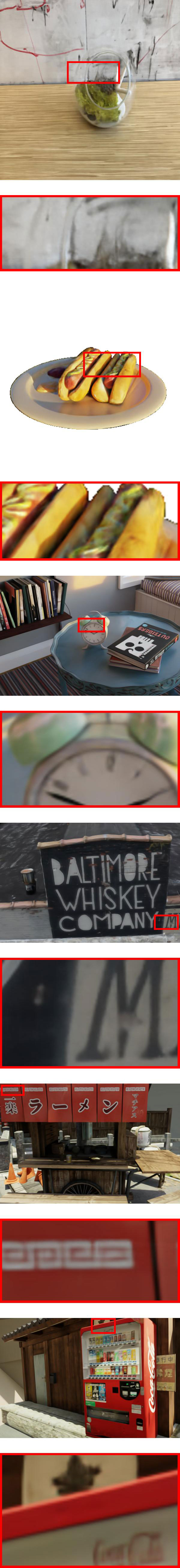}
  \end{minipage}
  ~
  \begin{minipage}[c]{0.14\linewidth}
  \includegraphics[width=1\linewidth]{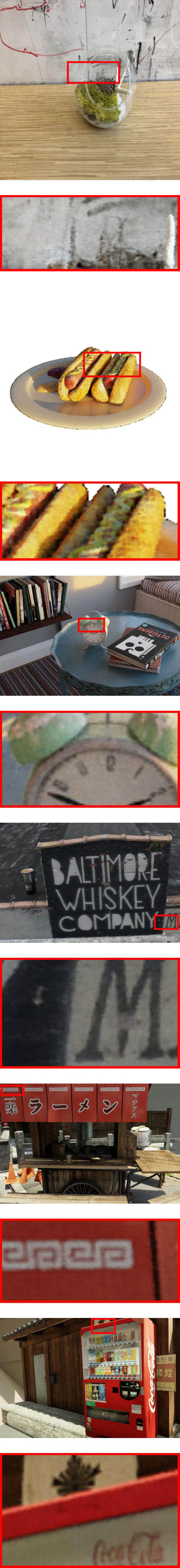}
  \end{minipage}
  ~
  \begin{minipage}[c]{0.14\linewidth}
  \includegraphics[width=1\linewidth]{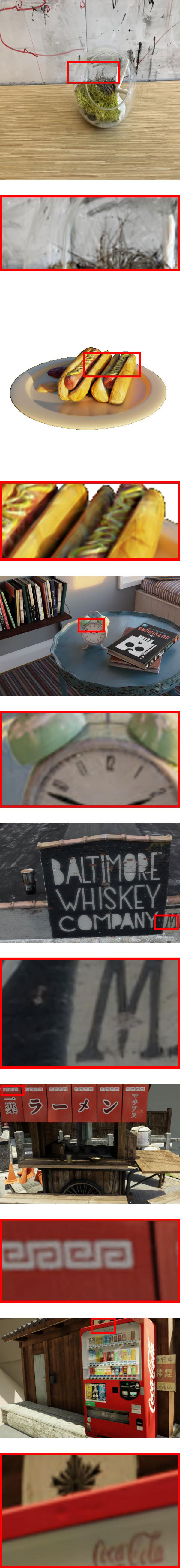}
  \end{minipage}
  ~
  \begin{minipage}[c]{0.14\linewidth}
  \includegraphics[width=1\linewidth]{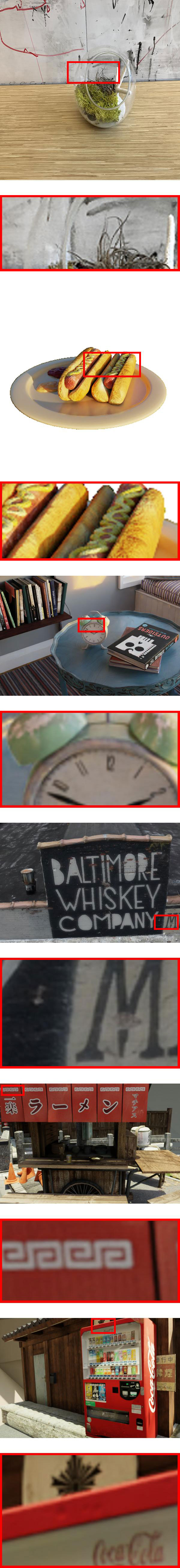}
  \end{minipage}
  
\end{minipage}
\end{minipage}
\\

   \caption{\textbf{Qualitative evaluations of our methods against naive two-stage baselines.} We compared the quality of synthesized novel-view images from our methods against that of vanilla 3DGS from SOTA methods. Top to bottom shows different scenes. The qualitative comparisons demonstrate that our methods outperforms existing approaches.
   }
   \label{fig5}
\end{figure*}

Table \ref{table II} summarizes the quantitative comparisons for novel-view image synthesis. Our SCINeRF achieves more than 33dB in the average PSNR, more than 0.95 in the average SSIM, and a mean LPIPS score of 0.04, surpassing the best SOTA + NeRF-based approach by $>$ 1.8dB in the PSNR. As illustrated in Fig. \ref{fig5}, the two-stage baseline methods frequently generate artifacts, such as blurring and fog-like effects, particularly evident in areas requiring fine details. These artifacts likely arise from the inherent limitations of SOTA methods, as their reconstructed images lack the consistent appearance across viewing angles, which is required for accurate NeRF and 3DGS-based scene representations. Such inconsistencies contaminate the scene modeling, resulting in visual distortions. In contrast, our end-to-end framework integrates camera pose optimization with scene representation, enabling high-quality reconstruction of fine details and consistent scene appearance across multiple views. This highlights the importance of jointly optimizing camera poses and scene representations for accurate and visually coherent novel-view synthesis.

Furthermore, we evaluate the computational efficiency of different SCI reconstruction algorithms, presented in Table \ref{table III}. For this analysis, we compared the training time (if applicable) and inference speeds of various methods. Notably, SOTA approaches return 2D frames directly from SCI images, whereas our methods first reconstruct a 3D scene representation and subsequently render 2D images. So, in this paper, we compare the inference speed of SOTA methods with the rendering speed (measured in FPS) for our methods.

Existing SOTA methods generally require extensive training time and cannot achieve real-time inference. For instance, EfficientSCI, the top-performing prior method, requires more than 100 hours (nearly five days) to train its Transformer-based model, and its inference speed is limited to only 2.6 FPS. Although our SCINeRF significantly outperforms the SOTA methods in image and scene representation quality, its computational demand remains high. The SCINeRF training process takes approximately 5 hours, and the rendering speed is just at 0.25 FPS, falling short of real-time requirements. These limitations are attributed to the reliance of SCINeRF on implicit scene representation and the computationally intensive volumetric rendering process inherited from NeRF. 

\begin{table*}
        \caption{\textbf{Quantitative comparisons of training time (in hrs, if applicable) and inference/rendering speed (FPS) of different methods on synthetic dataset.} Here we compare the inference speed of SOTA methods against the rendering speed of our proposed method since our SCINeRF and SCISplat need to render compressed images from estimated scenes. Our SCISplat prevails on both training time and inference speed by reading $<$1 hrs in training and 205 FPS in rendering.}
	\setlength\tabcolsep{8pt}
	\parbox{\textwidth}{
		\resizebox{\linewidth}{!}{
		\begin{tabular}{c|c|c|c|c|c|c}		
			\hline
            Methods & GAP-TV & PnP-FFDNet & PnP-FastDVDNet & EfficientSCI & SCINeRF & SCISplat \\
			
			\hline
			
			Training Time (hrs) $\downarrow$ &N/A &N/A &N/A &$\approx$100 hrs &{\underline{$\approx$5 hrs}} &{\textbf{$<$1hrs}} \\
			Inference Speed (FPS) $\uparrow$ &0.13 &0.01 &0.01 &{\underline{2.6}} &0.25 &{\textbf{205}} \\
            \hline		
	\end{tabular}}
    
	\label{table III}}

\end{table*}

\begin{table*}
        \caption{\textbf{Pose estimation performance of our SCINeRF and SCISplat on synthetic dataset.} The results are in the ATE metric. SCINeRF employs a linear trajectory representation, which achieve lower error in scenes with linear camera motion, (e.g., \textit{Cozy2room, Tanabata and Vender}), while the SCISplat, which optimize poses independently, fit better to complex trajectories, for example, snake-shape trajectories in \textit{Airplants} and arc-shape trajectories in \textit{Hotdog}.}
	\setlength\tabcolsep{18pt}
	\parbox{\textwidth}{
		\resizebox{\linewidth}{!}{
		\begin{tabular}{c|c|c|c|c|c|c}
			
			\hline
			ATE $\downarrow$ & Airplants & Hotdog & Cozy2room & Tanabata & Factory & Vender\\
			
			\hline
			
			SCINeRF &0.00502 &0.02068 &{\textbf{0.00015}} &{\textbf{0.00016}} &0.00065 &{\textbf{0.00023}} \\
			SCISplat &{\textbf{0.00459}} &{\textbf{0.01536}} &0.02028 &0.05117 &{\textbf{0.00059}} &0.00231 \\
			
            \hline
			
	\end{tabular}}

	\label{table IV}}
\end{table*}

To address these challenges, we incorporate the 3DGS framework into our SCISplat approach, significantly enhancing efficiency. SCISplat employs an explicit scene representation and a fast tile-base differentiable Gaussian rasterization process, accelerating training and rendering. As a result, SCISplat completes the training within 1 hour, a 5$\times$ improvement over SCINeRF, and achieves a rendering speed exceeding 200 FPS, a 820$\times$ speedup compared to SCINeRF. This efficiency makes SCISplat capable of rendering reconstructed 3D scenes from SCI images in real-time, a critical milestone for practical deployment in time-sensitive applications. The substantial speedup gives the advantage of integrating efficient computational frameworks while maintaining the high-quality scene representation of our method. 

As our method integrates 3D scene reconstruction with camera pose estimation during the training process, we further evaluate its performance in camera pose optimization. To achieve this, we calculate the absolute trajectory error (ATE) metric, which quantifies the deviation between the camera trajectories estimated by our methods and the ground truth trajectories. Table \ref{table IV} provides a detailed comparison of the ATE values of our SCINeRF and SCISplat approaches in different data sets. The results indicate that both SCINeRF and SCISplat deliver accurate camera pose estimations, with ATE values demonstrating a close match to the ground-truth trajectories.  Specifically, SCINeRF achieves consistently low ATE on datasets where ground truth trajectories adhere to linear, such as the datasets \textit{Cozy2room, Tanabata} and \textit{Factory} datasets. These linear trajectories align well with SCINeRF's design, which assumes a linear motion model for pose estimation, as indicated in Eq. \ref{linear_pose_assumption}.
On the other hand, SCISplat demonstrates superior adaptability to more complex and non-linear camera motions. For instance, in datasets like \textit{Airplants} (with snake-shaped trajectories) and \textit{Hotdog} (with arc-shaped trajectories), SCISplat outperforms SCINeRF in ATE, reflecting its ability to optimize camera poses individually without being constrained by linear trajectory assumptions. This flexibility is a significant advantage in real-world scenarios, where camera movements are often irregular or follow non-linear paths. This capability verifies the robustness of our method in simultaneously reconstructing 3D scenes and optimizing camera poses directly from SCI measurements. 

\subsection{Real Dataset}
\begin{figure*}[!tbp]

\begin{minipage}[c]{1.0\textwidth}
\begin{minipage}[c]{\linewidth}
\centering
  \begin{minipage}[c]{0.15\linewidth}
  \centering
  \small
  \ \\ \ Measurement
  \end{minipage}
  ~
  \begin{minipage}[c]{0.15\linewidth}
  \centering
  \small
  \ \\ PnP-FFDNet
  \end{minipage}
  ~
  \begin{minipage}[c]{0.15\linewidth}
  \centering
  \small
  \ \\ \ EfficientSCI
  \end{minipage}
  ~
  \begin{minipage}[c]{0.15\linewidth}
  \centering
  \small
  \ \\ \ SCINeRF
  \end{minipage}
  ~
  \begin{minipage}[c]{0.15\linewidth}
  \centering
  \small
  \ \\ \ SCISplat
  \end{minipage}
  ~
  \begin{minipage}[c]{0.15\linewidth}
  \centering
  \small
  \ \\ \ Original Scene
  \end{minipage}
\end{minipage}
\\
\begin{minipage}[c]{\linewidth}
\centering
  \begin{minipage}[c]{0.15\linewidth}
  \includegraphics[width=1\linewidth]{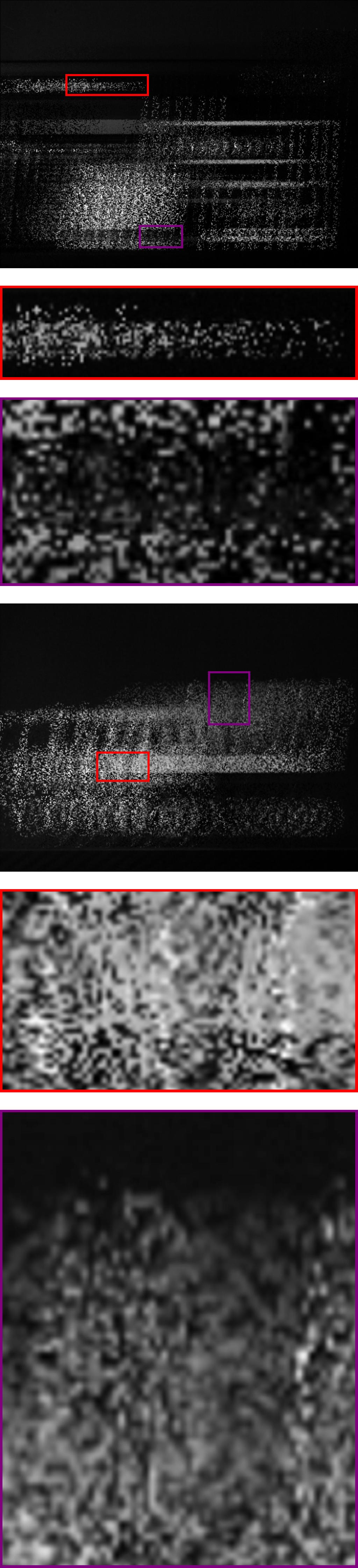}
  
  \end{minipage}
  ~
  \begin{minipage}[c]{0.15\linewidth}
  \includegraphics[width=1\linewidth]{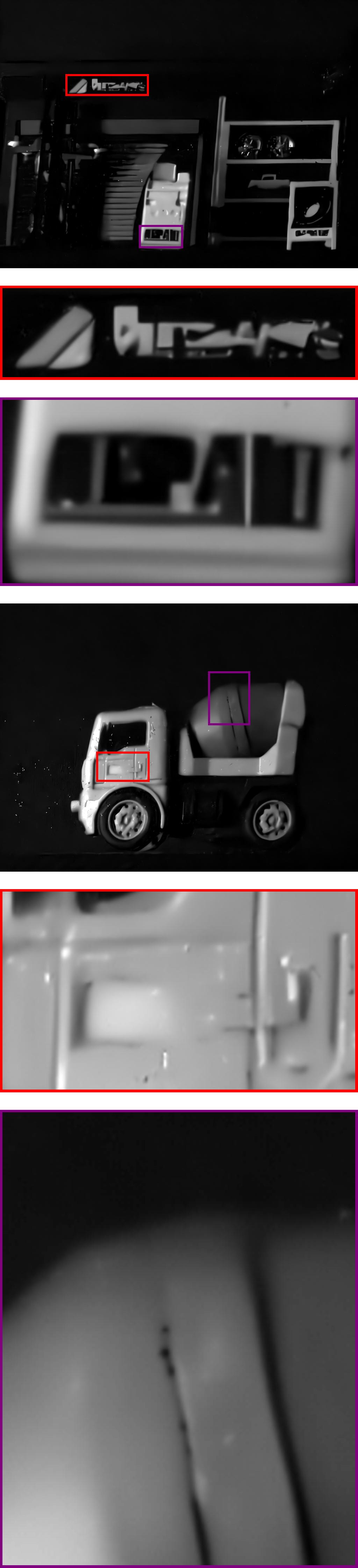}
  \end{minipage}
  ~
  \begin{minipage}[c]{0.15\linewidth}
  \includegraphics[width=1\linewidth]{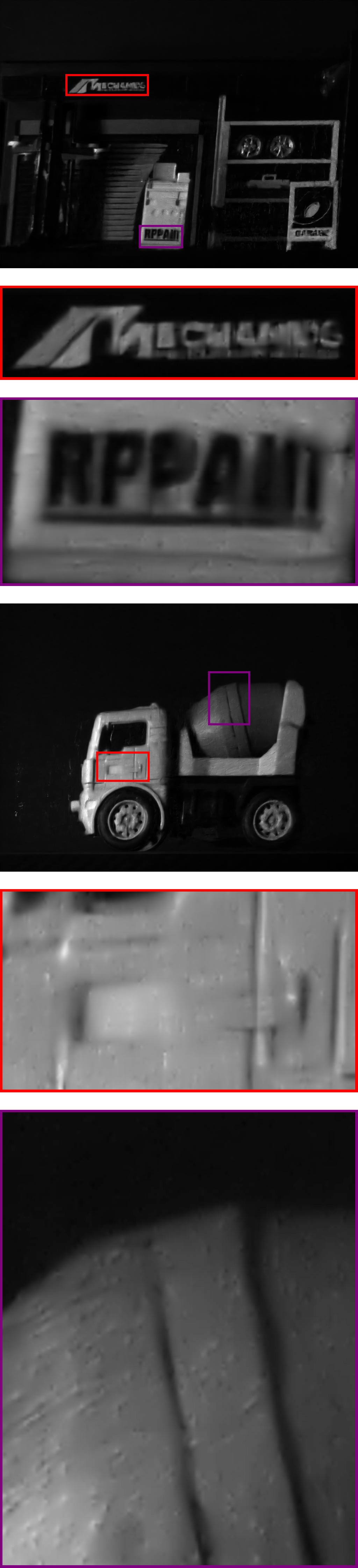}
  \end{minipage}
  ~
  \begin{minipage}[c]{0.15\linewidth}
  \includegraphics[width=1\linewidth]{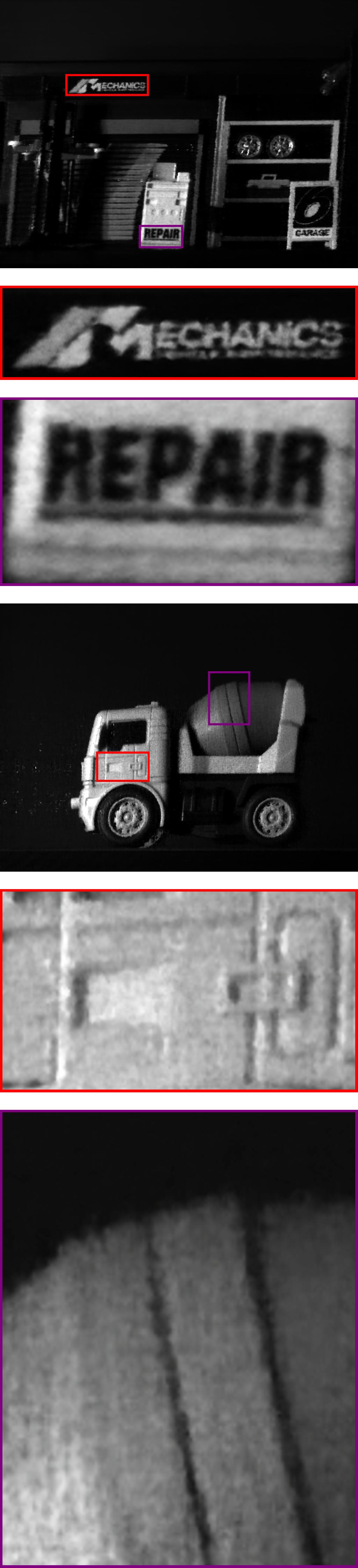}
  \end{minipage}
  ~
  \begin{minipage}[c]{0.15\linewidth}
  \includegraphics[width=1\linewidth]{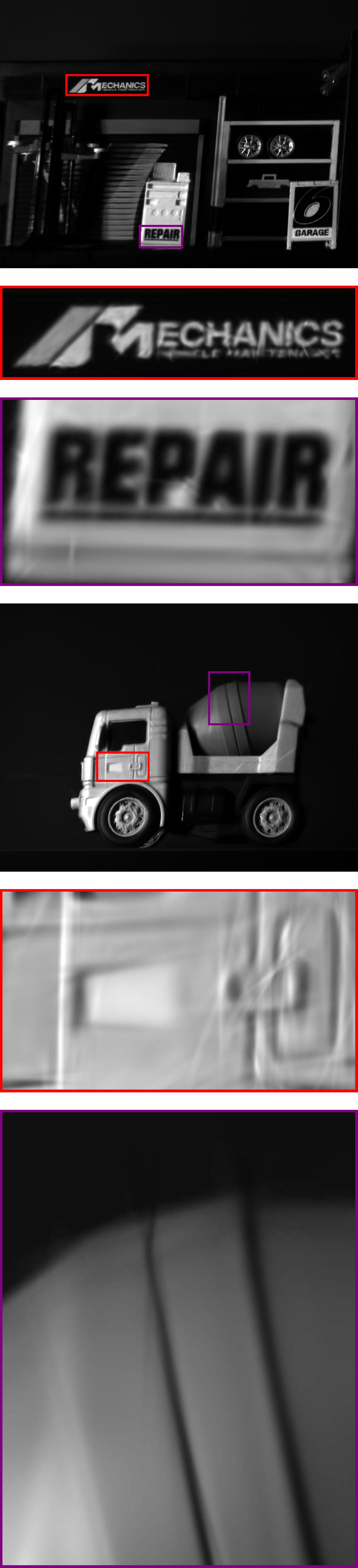}
  \end{minipage}
  ~
  \begin{minipage}[c]{0.15\linewidth}
  \includegraphics[width=1\linewidth]{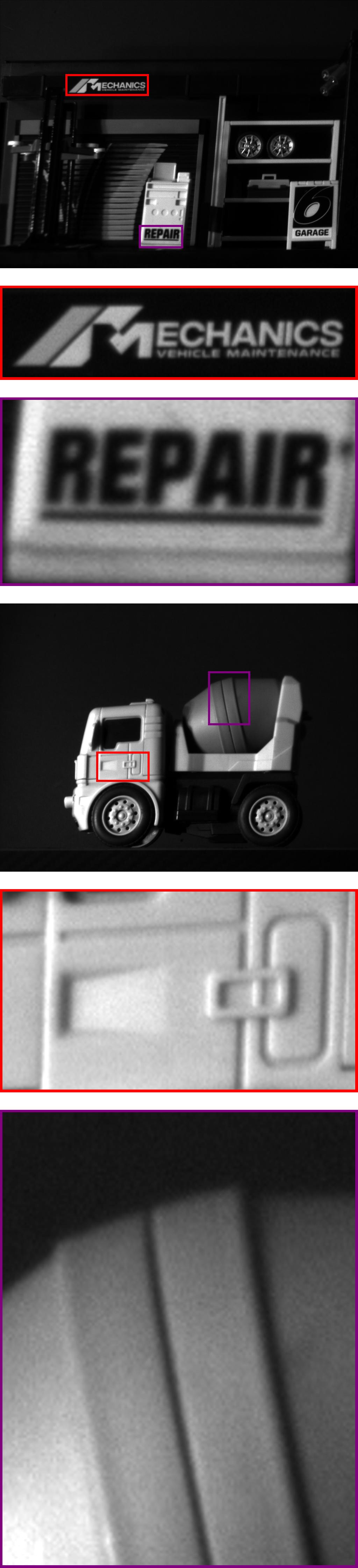}
  \end{minipage}
  
\end{minipage}
\end{minipage}
\\

   \caption{\textbf{Qualitative evaluations of our methods against SOTA SCI image restoration methods on real grayscale dataset captured by our system.} Top to bottom shows the results for different scenes. Since the compressed pixel-wise aligned ground truth images are unavailable, we capture separate scene images after capturing the snapshot compressed image used for reference. For qualitative evaluation purpose, we render images from the learned 3D scene representations by our methods. The experimental results show that our method can recover high-quality scene details in real datasets. Moreover, our SCISplat surpasses prior SOTA methods by recovering intricate details and avoiding high-frequency ``granular" noises.
   }
   \label{fig6}
\end{figure*}
To comprehensively evaluate the performance of our proposed methods on real-world datasets collected by our SCI imaging system in Fig. \ref{experimental setup}, we compare them against SOTA methods. Unlike synthetic datasets, where measurement noise is typically omitted for simplification, real datasets inherently contain noise, introducing two primary challenges. First, the modulation masks deviate from ideal binary patterns,  which means that the mask values on each pixel location are no longer 0 or 1. Second, the modulated and aggregated pixel intensities in input frames may differ from the actual measurement values recorded by the camera sensor due to the presence of noise, as described in Eq. \eqref{eq_formation}. These discrepancies significantly complicate the SCI image reconstruction process.

The impact of these challenges is illustrated in Fig. \ref{fig6}, which presents qualitative results for real grayscale datasets. It is important to note that, due to the lack of pixel-wise aligned ground-truth images in real datasets, we captured separate scene images after recording the SCI measurements for reference purposes. The results demonstrate that existing SOTA methods face considerable limitations when applied to real datasets. For example, PnP-FFDNet can reconstruct only the general shape of the scenes but fails to recover fine details, highlighting its sensitivity to noise. EfficientSCI performs moderately better by retrieving more scene details. However, similar to the results in the synthetic dataset, it suffers from a pronounced generalization gap. This manifests as blurred or distorted details in reconstructed images, leading to noticeable artifacts and diminished image quality. 

In contrast, our SCINeRF framework exhibits significant improvements in the handling of real data sets. Using its advanced 3D scene representation capabilities, SCINeRF effectively recovers scenes with intricate details and achieves superior qualitative performance compared to SOTA methods. However, SCINeRF has a relatively low noise suppression performance, which introduces high-frequency ``granular" noise in the rendered images, which may affect the perceptual quality. 

\begin{figure*}[!tbp]

\begin{minipage}[c]{1.0\textwidth}
\begin{minipage}[c]{\linewidth}
\centering
  \begin{minipage}[c]{0.15\linewidth}
  \centering
  \small
  \ \\ \ Measurement
  \end{minipage}
  ~
  \begin{minipage}[c]{0.15\linewidth}
  \centering
  \small
  \ \\ PnP-FFDNet
  \end{minipage}
  ~
  \begin{minipage}[c]{0.15\linewidth}
  \centering
  \small
  \ \\ \ EfficientSCI
  \end{minipage}
  ~
  \begin{minipage}[c]{0.15\linewidth}
  \centering
  \small
  \ \\ \ SCINeRF
  \end{minipage}
  ~
  \begin{minipage}[c]{0.15\linewidth}
  \centering
  \small
  \ \\ \ SCISplat
  \end{minipage}
  ~
  \begin{minipage}[c]{0.15\linewidth}
  \centering
  \small
  \ \\ \ Original Scene
  \end{minipage}
\end{minipage}
\\
\begin{minipage}[c]{\linewidth}
\centering
  \begin{minipage}[c]{0.15\linewidth}
  \includegraphics[width=1\linewidth]{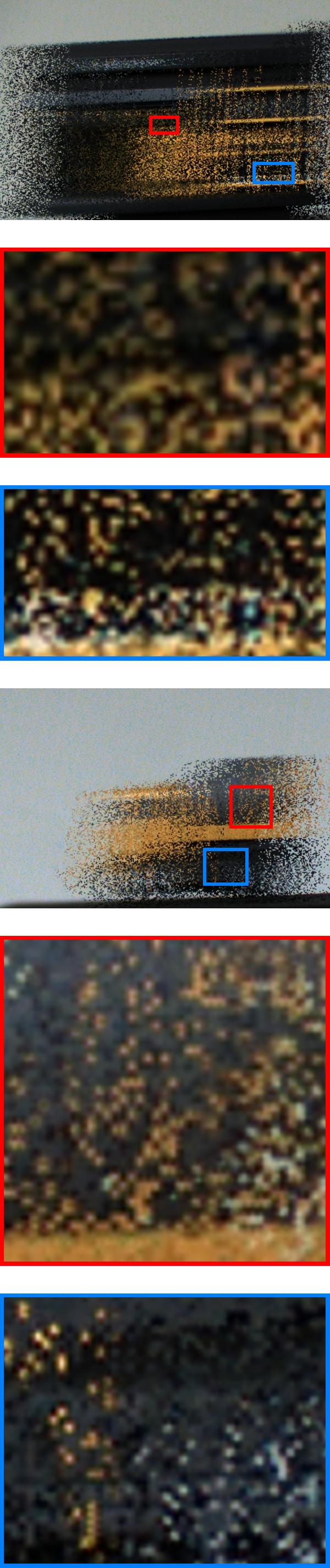}
  
  \end{minipage}
  ~
  \begin{minipage}[c]{0.15\linewidth}
  \includegraphics[width=1\linewidth]{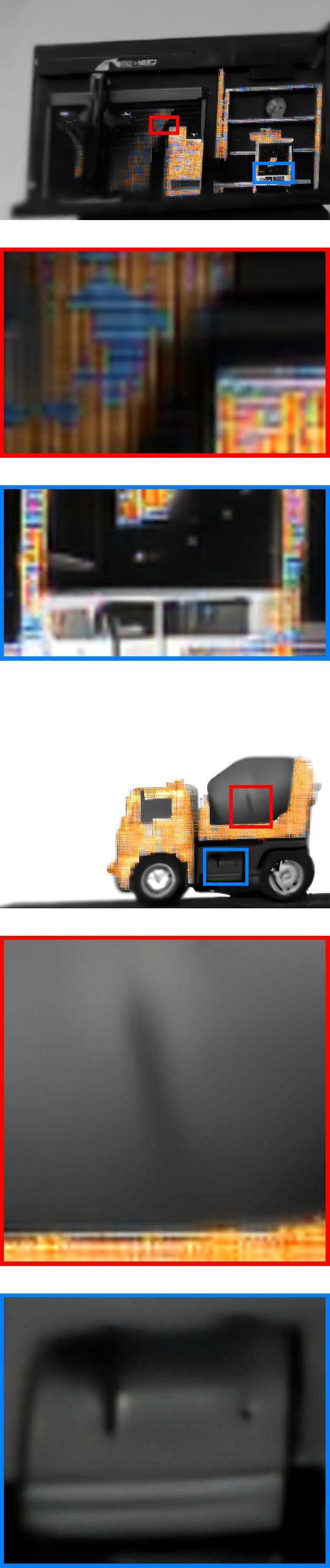}
  \end{minipage}
  ~
  \begin{minipage}[c]{0.15\linewidth}
  \includegraphics[width=1\linewidth]{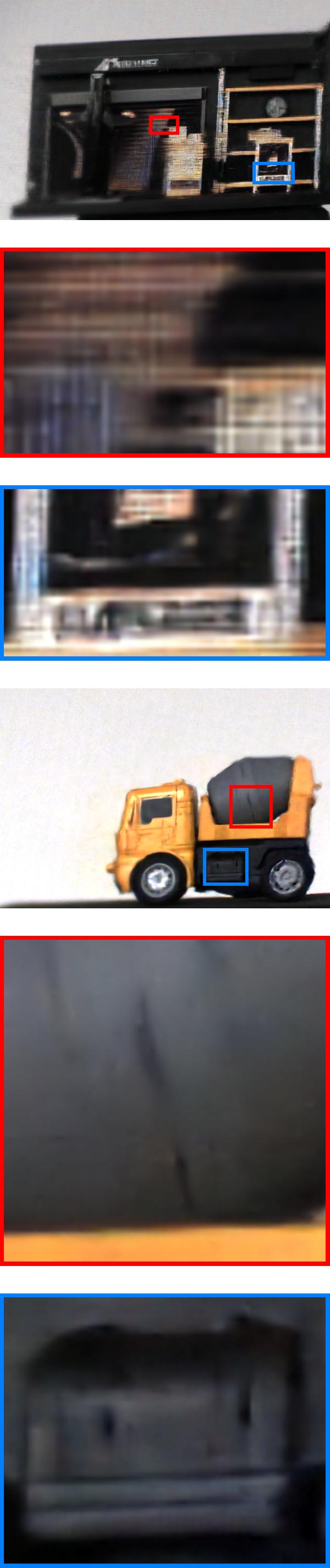}
  \end{minipage}
  ~
  \begin{minipage}[c]{0.15\linewidth}
  \includegraphics[width=1\linewidth]{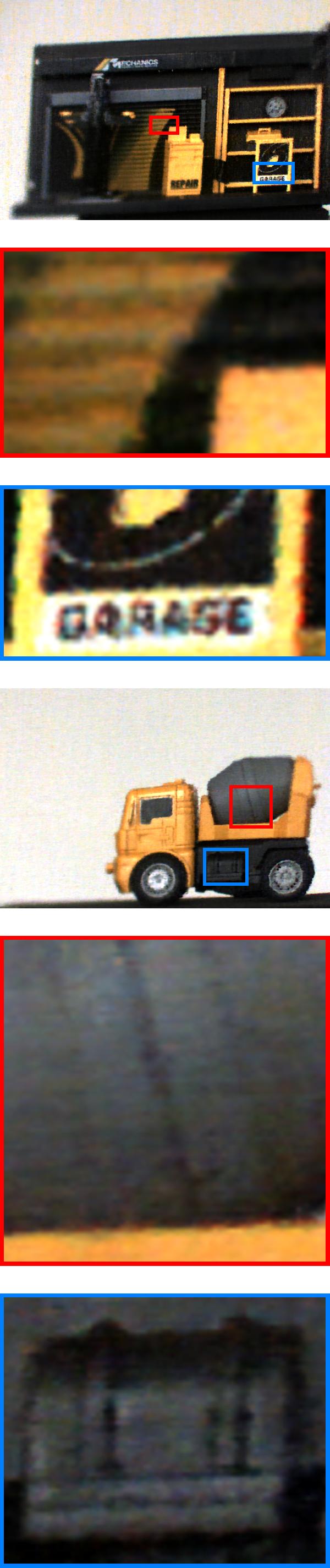}
  \end{minipage}
  ~
  \begin{minipage}[c]{0.15\linewidth}
  \includegraphics[width=1\linewidth]{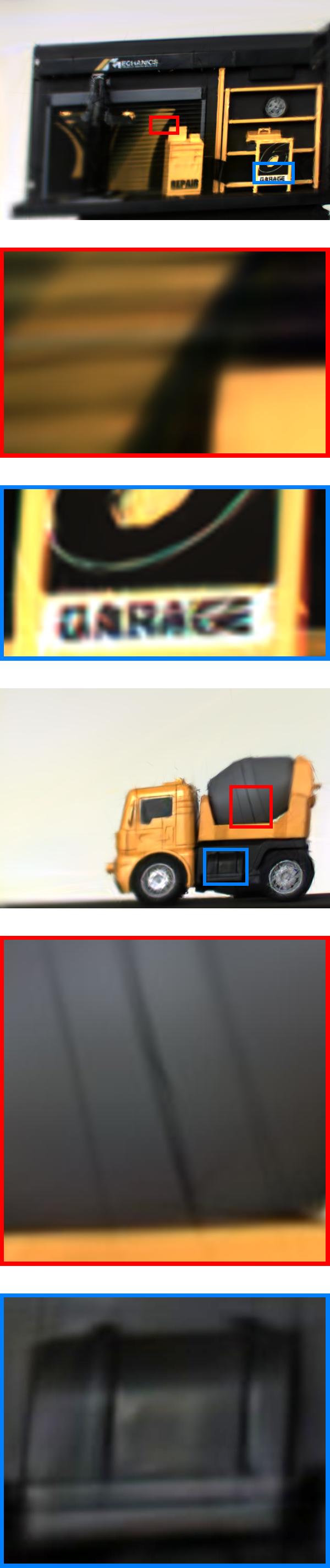}
  \end{minipage}
  ~
  \begin{minipage}[c]{0.15\linewidth}
  \includegraphics[width=1\linewidth]{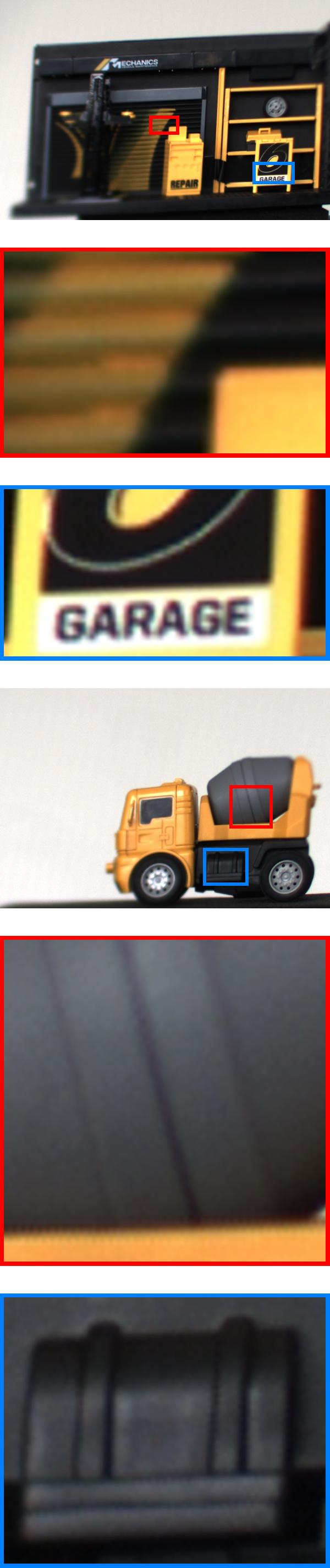}
  \end{minipage}
  
\end{minipage}
\end{minipage}
\\

   \caption{\textbf{Qualitative evaluations of our methods against SOTA SCI image restoration methods with real color dataset.} Top to bottom shows different scenes. Due to the generalization gap, the prior SOTA methods experience significant performance degradation on real dataset, leading to unsatisfactory scene details retrieved from the real color SCI image. Our methods, however, keep high performance on real color data by successfully recovering compressed multi-view images with rich details. 
   }
   \label{fig7}
\end{figure*}

To address this limitation and enhance noise suppression capabilities, we apply our SCISplat framework to the same real datasets. As depicted in Fig. \ref{fig6}, SCISplat not only retains SCINeRF's ability to recover fine details but also effectively eliminates high-frequency noise, resulting in significantly improved output quality. Furthermore, Fig. \ref{fig7} shows the performance of different SCI reconstruction methods on real color datasets. The results reveal that previous SOTA methods suffer severe degradation in image quality when applied to real color datasets, failing to produce a satisfactory reconstruction. In contrast, our SCINeRF and SCISplat maintain consistently high performance by successfully recovering compressed images with great detail and reduced noise artifacts. These findings demonstrate the robustness and adaptability of our approach to tackle the challenges posed by real-world SCI data, which strongly supports the subsequent deployment of our proposed method.   

\section{Ablation Study}
\subsection{Mask overlapping rate}

We evaluated the effect of different mask-overlapping rates when modulating the input frames during the SCI image formation process. During the imaging process, the camera moves along a trajectory, and a sequence of 2D masks is applied to modulate input multi-view images, generating a compressed measurement as described in Eq. \eqref{eq_formation}. It is possible that there is a spatial overlapping between masks. For a given number $N$ masks throughout the exposure time, the probability of assigning the value of 1 to a specific pixel location is constant, and we define it as the mask overlapping rate (OR). The OR is mathematically expressed as follows:
\begin{equation}\label{eq_overlapping_rate}
\mathbf{OR}(x,y) = \frac{\sum_{i=1}^{N}\mathbf{M}_i(x,y)}{N},     
\end{equation}
where $\mathbf{OR}$ denotes mask overlapping rate, $\mathbf{M}_i$ is the $i$-th mask and $(x,y)$ indicates pixel coordinate, and $N$ is total number of compressed images. From Eq.~\eqref{eq_overlapping_rate}, it becomes evident that a lower mask overlapping rate results in fewer pixels being assigned a value of 1 in the spatial domain. Consequently, the mask becomes sparse, leading to a lower sampling rate and a higher degree of information loss. Conversely, a higher overlapping rate increases the number of pixels modulated across frames, introducing ambiguity in pixel values and potential deterioration in the quality of the estimated scenes and rendered images. Therefore, selecting an appropriate mask overlapping rate involves a trade-off between minimizing information loss and mitigating pixel ambiguity, making the appropriate mask overlapping rate neither too high nor too low. 

\begin{table}[]
     \caption{\textbf{Ablation studies on the effect of mask overlapping rate.} The results demonstrate that the image quality first increases and then decreases when the mask overlapping rate increases. The best mask overlapping rate is 0.25.}
     \resizebox{1.0\linewidth}{!}{
    \footnotesize
    \setlength\tabcolsep{4pt}
    \begin{tabular}{c|ccc}
     \hline
     & \multicolumn{3}{c}{\scriptsize{ Synthetic Dataset}} \\
     \hline
      Mask Overlapping Rate & \scriptsize PSNR$\uparrow$ &\scriptsize SSIM$\uparrow$ & \scriptsize LPIPS$\downarrow$\\
 	\hline
    \scriptsize 0.125 & \scriptsize 31.93 & \scriptsize 0.956 & \scriptsize 0.052 \\
 	 \scriptsize 0.25 & \scriptsize {\textbf{33.61}} & \scriptsize {\textbf{0.959}} & \scriptsize {\textbf{0.042}} \\
     \scriptsize 0.5 & \scriptsize 32.92 & \scriptsize 0.949 & \scriptsize 0.046 \\
     \scriptsize 0.75 & \scriptsize 30.08 & \scriptsize 0.891 & \scriptsize 0.139 \\
			
     \hline
    \end{tabular}
     }
     \label{table_mask}
\end{table}

We conducted experiments using synthetic datasets to validate this hypothesis and empirically identify the optimal overlapping rate. We construct multiple SCI measurements by modulating the same set of input images with masks, the overlapping rates of which range from 0.125 to 0.75. The evaluation results, summarized in Table 5, show that when the overlapping rate increases from 0.125 to 0.25, the image quality improves. This improvement reflects the advantages of sampling more pixels, which enhances the richness of captured information. However, when the overlapping rate is increases beyond 0.25, the image quality deteriorates, which is attributed to the growing pixel ambiguity. 
This phenomenon is particularly evident in extreme cases. For instance, when the overlapping rate reaches 1, the SCI image formation model degenerates to that of a blurred image. Recovering sharp images or sharp 3D scenes from such blurred measurements is severely ill-posed. Based on these findings, we empirically select an overlapping rate of 0.25 as the upper limit of all the experiments. 



\subsection{Interpolation threshold}

\begin{figure*}[!htbp]
  \centering
   \vspace{1em}
    \includegraphics[width=0.99\linewidth]{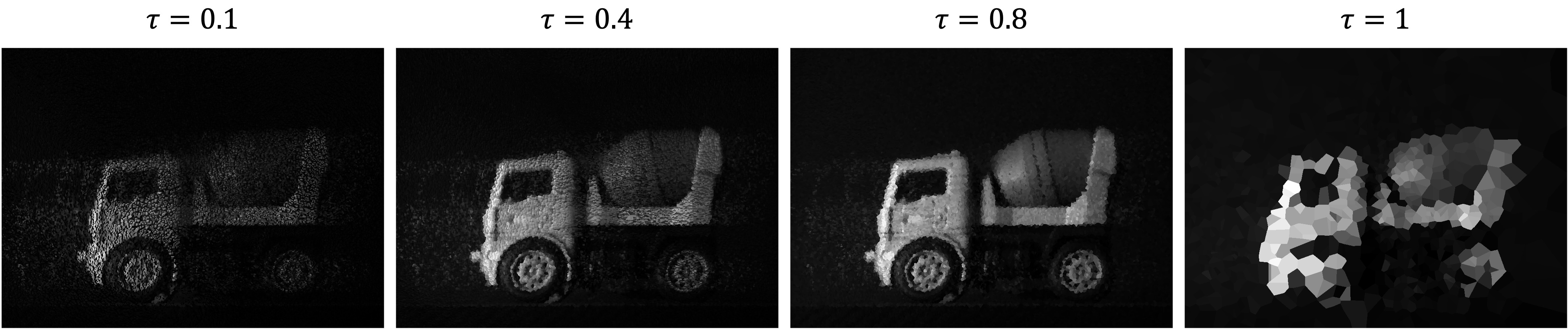}
    \caption{\textbf{Ablation studies on different thresholds $\tau$ for interpolating real data.} At a low threshold $\tau=0.1$, the interpolated image is overwhelmed by measurement noise. At a high threshold $\tau=1$, the image retains only a vague outline of the captured object. Therefore, we select $\tau=0.8$ as a balanced compromise.}
    \label{fig10}
\end{figure*}

We evaluated the effect of different interpolation thresholds $\tau$ in our SCISplat framework. A higher threshold effectively suppresses noise but makes fewer retained pixels, leading to more information loss. A lower threshold, on the other hand, retains more pixels while allowing excessive noise to present in the interpolated frames. We perform experiments with different threshold values on the real data set, as shown in Fig. \ref{fig10}. With a low threshold of 0.1, the interpolated frames are overwhelmed with measurement noise, contaminating useful scene details.
In contrast, a high threshold, like 1, results in frames with minimal noise but only preserves a vague outline of the reconstruction scene. Based on these empirical observations, we select a threshold of 0.8 for real datasets.

\begin{table}[]
     \caption{\textbf{Ablation studies on different initialization methods.} Both random initialization and COLMAP fail to retrieve any useful details from the degraded frame and thus cannot be applied in our initialization protocol. DUSt3R gives dense points but inaccurate poses, which affects the performance. VGGSfM, which is employed by our SCISplat pipeline, consistently gives relatively reliable point clouds and relatively accurate camera poses, reflected by significantly better reconstruction quality.}
     \resizebox{1.0\linewidth}{!}{
    \footnotesize
    \setlength\tabcolsep{8pt}
    \begin{tabular}{c|ccc}
     \hline
     & \multicolumn{3}{c}{\scriptsize{ Synthetic Dataset}} \\
     \hline
      Initialization & \scriptsize PSNR$\uparrow$ &\scriptsize SSIM$\uparrow$ & \scriptsize LPIPS$\downarrow$\\
 	\hline
    \scriptsize Random Initialization & \scriptsize N/A & \scriptsize N/A & \scriptsize N/A \\
 	 \scriptsize COLMAP & \scriptsize N/A & \scriptsize N/A & \scriptsize N/A \\
     \scriptsize DUSt3R & \scriptsize 32.12 & \scriptsize 0.940 & \scriptsize 0.052 \\
     \scriptsize VGGSfM & \scriptsize {\textbf{35.94}} & \scriptsize {\textbf{0.977}} & \scriptsize {\textbf{0.014}} \\
			
     \hline
    \end{tabular}
     }
     \label{table_threshold}
\end{table}

\subsection{Initialization method} 

We conducted extensive experiments on synthetic datasets using various initialization methods to setup SCISplat training. Initially, we implemented a random initialization approach, where a random point cloud is generated within a predefined bounding box, and camera poses are initialized following a linear trajectory. This method fails to provide meaningful results due to the lack of valid scene information, which hinders the 3D geometric structures and makes it too difficult for 3DGS to optimize. Subsequently, we evaluate the performance of conventional SfM software COLMAP \cite{schonberger2016structure}, which is widely utilized in 3DGS-based methods. Using COLMAP on interpolated degraded framed generated via the process described in Eq. \eqref{degraded_frame2} also fails to produce valid point clouds or accurate camera poses, which is not suitable for the initialization of SCISplat. We further tested DUSt3R \cite{wang2024dust3r}, a Transformer-based dense stereo model employed in some of the latest 3DGS-based frameworks. While DUSt3R demonstrates the ability to generate dense point clouds from sparse-view images, it returns inaccurate camera poses when applied to degraded frames, affecting the reconstruction quality.
In contrast, VGGSfM \cite{wang2024vggsfm}, which is applied by our SCISplat framework, delivers reliable point clouds and relatively accurate camera poses. VGGSfM provides good initial values and enables the SCISplat framework to achieve superior reconstruction performance. Quantitative comparisons in Table \ref{table_threshold} show the effectiveness of the selected initialization method. Both random initialization and COLMAP fail to yield valid results, and VGGSfM applied in our framework outperforms DUSt3R by over 3.8dB in PSNR.

\subsection{Number of initial points}
To evaluate the impact of down-sampling operation within the initialization protocol on final rendering quality, we conduct experiments on two synthetic datasets with a relatively high number of initial points: \textit{Airplants}, with 15519 initial points, and \textit{Factory}, with 29249 initial points. These scenes come with different camera trajectories: \textit{Airplants} follows a snake-shape trajectory, and \textit{Factory} exhibits a relatively linear trajectory. Table \ref{table_downsample} presents the results of training our SCISplat model with various numbers of points, including using the entire set of points without downsampling and points downsampled to 100,000, 5000 and 1000. The results show that training with the full set of points, which may include additional scene details and more noise, generally leads to degraded rendering quality, particularly in the \textit{Airplants}. This can be attributed to the higher noise levels in the interpolated frames generated by the complex, snake-shaped camera trajectory. Conversely, the down-sampling operation appears less critical in the \textit{Factory} scene with simpler camera motions. The linear camera trajectory reduces noise in the degraded frames, which may lead to a cleaner point cloud. However, when we continue to down-sample points to 5000 and 1000, insufficient initial points negatively impact performance in both scenes. This limitation comes from the inability of the 3DGS densification strategy to effectively grow Gaussians from sparse point coverage, thereby limiting the model's capabilities to recover finer scene details. 

\begin{table}[]
     \caption{\textbf{Ablation studies on different downsample thresholds for initial points} We conduct qualitative experiments with different numbers of initial points on \textit{Airplants} and \textit{Factory} in synthetic dataset. The words ``\textit{w/o} downsample" indicate that no downsampling is applied to the initial point cloud. Subsequently, the downsample threshold $n$ is set to 10000, 5000, and 1000 to determine the optimal value.}
     \resizebox{1.0\linewidth}{!}{
    \footnotesize
    \setlength\tabcolsep{2pt}
    \begin{tabular}{c|ccc|ccc}
     \hline
     & \multicolumn{3}{c|}{Airplants} & \multicolumn{3}{c}{Factory} \\
      Threshold \textit{n} &  PSNR$\uparrow$ & SSIM$\uparrow$ &  LPIPS$\downarrow$ &  PSNR$\uparrow$ & SSIM$\uparrow$ &  LPIPS$\downarrow$\\
 	\hline
    \textit{w/o} downsample &  30.56 &  0.937 &  0.059 &  {\textbf{38.92}} &  {\textbf{0.975}} &  {\textbf{0.009}}\\
 	  10000 points &  {\textbf{31.45}} &  {\textbf{0.950}} &  {\textbf{0.036}} &  {\textbf{38.92}} &  {\textbf{0.975}} &  0.010\\
      5000 points &  31.36 &  0.948 &  0.040 &  38.51 &  0.974 &  0.010\\
      1000 points & 30.91 &  0.944 & 0.042 & 37.95 & 0.969 & 0.012\\
			
     \hline
    \end{tabular}
     }
     \label{table_downsample}
     \vspace{-1em}
\end{table}

\subsection{Densification strategy}

\begin{figure}[!tbp]
  \centering
    \includegraphics[width=0.99\linewidth]{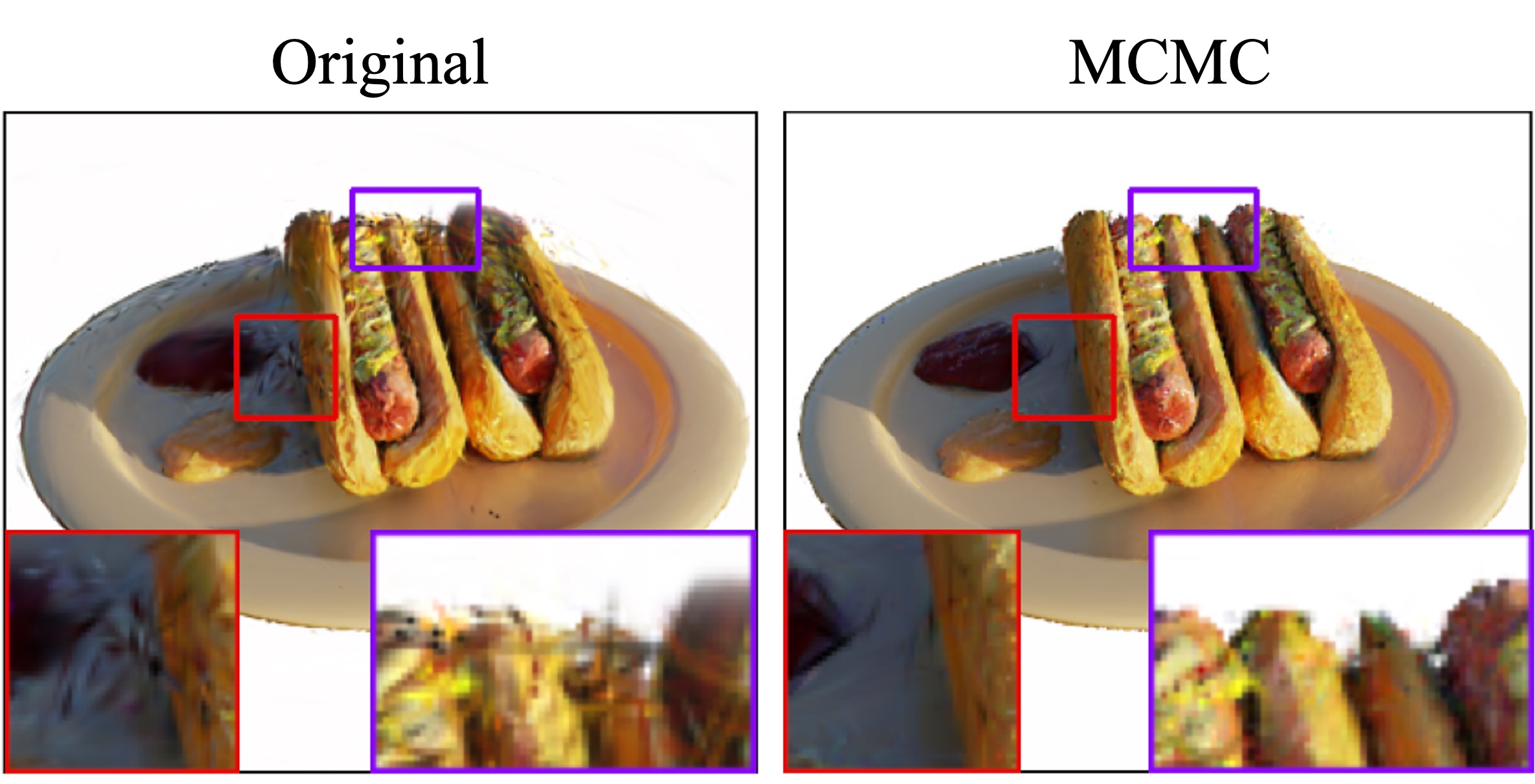}
    \caption{\textbf{Qualitative results with different densification strategies.} The original densification strategy in 3DGS introduces high opacity noise around the edges and causes noticeable deficiencies. On the other hand, our pipeline applies MCMC densification strategy in 3DGS training with superior image quality.}
    \label{fig11}
\end{figure}

As stated in previous sections, we improve the densification strategy of our SCISplat framework by replacing the original strategy with a more robust MCMC strategy to better align with the image formation model of SCI and to avoid the risk of converging into the sub-optimal solutions during the training of Gaussians. The results presented in Table~\ref{table_downsample} demonstrate the significant improvements of the new MCMC densification strategy over the original one. Our model incorporating MCMC outperforms the traditional densification strategy on the synthetic dataset by approximately 2.2dB on average PSNR. Fig. \ref{fig11} further qualitatively highlights the improvement of the MCMC strategy over the original densification approach. Sticking to the original densification strategy results in numerous floaters and high-opacity artifacts, particularly around the edges of the objects. In contrast, the MCMC strategy produces much cleaner results with more stable and higher-quality reconstructions, from which finer details and smoother surfaces are more consistently preserved. 

\subsection{Pose optimization}

As shown in Table \ref{table8}, we evaluate the effect of pose optimization by comparing the quality of the rendered images of our entire SCISplat pipeline with that without pose optimization. In the latter case, camera poses come from the initialization protocol and remain fixed throughout the training process. The quantitative results reveal that omitting pose optimization results in significant degradation in image quality, with an average PSNR reduction of more than 1.5dB in the synthetic dataset. This performance drop can be attributed to the fact that the initial camera poses are derived from degraded frames computed using Eq. \eqref{degraded_frame1} and Eq. \eqref{degraded_frame2}. These degraded frames come with noise, therefore inevitably affecting the accuracy of the initialization module. The experimental results highlight that the initial camera poses obtained from noise-degraded frames are often sub-optimal. Without further refinement, these inaccuracies will deteriorate the rendered image quality. The results presented in Table \ref{table8} validate the necessity of pose optimization in our SCISplat framework. The joint-optimization process ensures a more robust and high-quality reconstruction by optimizing initial poses together with the represented 3D scene.  

\begin{table}[]
     \caption{\textbf{Ablation studies on densification strategy and pose optimization.} This table shows the average metrics on synthetic datasets, demonstrating the impact of enabling or disabling pose optimization during the training procedure. The results also qualitatively prove the importance of using the MCMC strategy with pose optimization to achieve optimal performance.}
     \resizebox{1.0\linewidth}{!}{
    \footnotesize
    \setlength\tabcolsep{2pt}
    \begin{tabular}{c|ccc|ccc}
     \hline
     & \multicolumn{3}{c|}{\textit{w/} pose optimization} & \multicolumn{3}{c}{\textit{w/o} pose optimization} \\
      Strategy &  PSNR$\uparrow$ & SSIM$\uparrow$ &  LPIPS$\downarrow$ &  PSNR$\uparrow$ & SSIM$\uparrow$ &  LPIPS$\downarrow$\\
 	\hline
    original &  33.73 &  0.965 &  0.029 &  32.54 &  0.956 &  0.042\\
 	  MCMC &  {\textbf{35.94}} &  {\textbf{0.977}} &  {\textbf{0.014}} &  34.43 &  0.971 &  0.013\\
      			
     \hline
    \end{tabular}
     }
     \label{table8}
\end{table}
\section{Conclusion}

In this paper, we first present SCINeRF, a novel approach for 3D scene representation learning from a single snapshot compressed image. SCINeRF exploits neural radiance fields as its underlying scene representation due to its impressive representation capability. The physical image formation process of an SCI image is exploited to formulate the training objective for joint NeRF training and camera pose optimization. Different from prior works, our method considers the underlying 3D scene structure to ensure multi-view consistency among recovered images. To further improve the scene representation quality and training/rendering speed, we introduce SCISplat, a 3DGS-based approach for efficiently reconstructing a 3D scene from a single SCI image. By leveraging popular 3D Gaussian Splatting as the core scene representation, SCISplat offers remarkable scene fidelity and exceptional speed in both training and rendering. To initiate the training of SCISplat, we propose a novel initialization protocol that robustly derives the initial point cloud and camera poses from a single SCI measurement. To validate the effectiveness of our methods, we conduct thorough evaluations against existing state-of-the-art techniques for SCI image reconstruction with both synthetic and real datasets. Extensive experimental results demonstrated the superior performance of our methods in comparison to existing methods, and the necessity to consider the underlying 3D scene structures for SCI decoding. Moreover, our SCISplat outperforms all prior methods in terms of training efficiency and rendering speed. In the future, we will explore the potential applications of our methods, such as capturing and recovering 3D structures of physical and biological reactions.




\bibliographystyle{IEEEtran}

\bibliography{reference}

\begin{thebibliography}{1}
\providecommand{\url}[1]{#1}
\csname url@samestyle\endcsname
\providecommand{\newblock}{\relax}
\providecommand{\bibinfo}[2]{#2}
\providecommand{\BIBentrySTDinterwordspacing}{\spaceskip=0pt\relax}
\providecommand{\BIBentryALTinterwordstretchfactor}{4}
\providecommand{\BIBentryALTinterwordspacing}{\spaceskip=\fontdimen2\font plus
\BIBentryALTinterwordstretchfactor\fontdimen3\font minus
  \fontdimen4\font\relax}
\providecommand{\BIBforeignlanguage}[2]{{%
\expandafter\ifx\csname l@#1\endcsname\relax
\typeout{** WARNING: IEEEtran.bst: No hyphenation pattern has been}%
\typeout{** loaded for the language `#1'. Using the pattern for}%
\typeout{** the default language instead.}%
\else
\language=\csname l@#1\endcsname
\fi
#2}}
\providecommand{\BIBdecl}{\relax}
\BIBdecl

\bibitem{li2024scinerf}
Y.~Li, X.~Wang, P.~Wang, X.~Yuan, and P.~Liu, ``{SCINeRF: Neural Radiance
  Fields from a Snapshot Compressive Image},'' in \emph{Proceedings of the
  IEEE/CVF Conference on Computer Vision and Pattern Recognition}, 2024, pp.
  10\,542--10\,552.

\end{thebibliography}


\begin{thebibliography}{10}
\providecommand{\url}[1]{#1}
\csname url@samestyle\endcsname
\providecommand{\newblock}{\relax}
\providecommand{\bibinfo}[2]{#2}
\providecommand{\BIBentrySTDinterwordspacing}{\spaceskip=0pt\relax}
\providecommand{\BIBentryALTinterwordstretchfactor}{4}
\providecommand{\BIBentryALTinterwordspacing}{\spaceskip=\fontdimen2\font plus
\BIBentryALTinterwordstretchfactor\fontdimen3\font minus
  \fontdimen4\font\relax}
\providecommand{\BIBforeignlanguage}[2]{{%
\expandafter\ifx\csname l@#1\endcsname\relax
\typeout{** WARNING: IEEEtran.bst: No hyphenation pattern has been}%
\typeout{** loaded for the language `#1'. Using the pattern for}%
\typeout{** the default language instead.}%
\else
\language=\csname l@#1\endcsname
\fi
#2}}
\providecommand{\BIBdecl}{\relax}
\BIBdecl

\bibitem{yuan2021snapshot}
X.~Yuan, D.~J. Brady, and A.~K. Katsaggelos, ``Snapshot compressive imaging:
  Theory, algorithms, and applications,'' \emph{IEEE Signal Processing
  Magazine}, vol.~38, no.~2, pp. 65--88, 2021.

\bibitem{Llull13_OE_CACTI}
P.~Llull, X.~Liao, X.~Yuan, J.~Yang, D.~Kittle, L.~Carin, G.~Sapiro, and D.~J.
  Brady, ``Coded aperture compressive temporal imaging,'' \emph{Opt. Express},
  vol.~21, no.~9, pp. 10\,526--10\,545, May 2013.

\bibitem{Candes2006TIT}
E.~J. Cand{\`e}s, J.~Romberg, and T.~Tao, ``Robust uncertainty principles:
  Exact signal reconstruction from highly incomplete frequency information,''
  \emph{IEEE Transactions on information theory}, vol.~52, no.~2, pp. 489--509,
  2006.

\bibitem{Donoho2006TIT}
D.~L. Donoho, ``Compressed sensing,'' \emph{IEEE Transactions on information
  theory}, vol.~52, no.~4, pp. 1289--1306, 2006.

\bibitem{yuan2015generalized}
X.~Yuan, ``Generalized alternating projection based total variation
  minimization for compressive sensing,'' in \emph{2016 IEEE International
  conference on image processing (ICIP)}.\hskip 1em plus 0.5em minus
  0.4em\relax IEEE, 2016, pp. 2539--2543.

\bibitem{liao2014generalized}
X.~Liao, H.~Li, and L.~Carin, ``Generalized alternating projection for
  weighted-2,1 minimization with applications to model-based compressive
  sensing,'' \emph{SIAM Journal on Imaging Sciences}, vol.~7, no.~2, pp.
  797--823, 2014.

\bibitem{liu2018rank}
Y.~Liu, X.~Yuan, J.~Suo, D.~J. Brady, and Q.~Dai, ``Rank minimization for
  snapshot compressive imaging,'' \emph{IEEE transactions on pattern analysis
  and machine intelligence}, vol.~41, no.~12, pp. 2990--3006, 2018.

\bibitem{qiao2020deep}
M.~Qiao, Z.~Meng, J.~Ma, and X.~Yuan, ``Deep learning for video compressive
  sensing,'' \emph{Apl Photonics}, vol.~5, no.~3, 2020.

\bibitem{cheng2020birnat}
Z.~Cheng, R.~Lu, Z.~Wang, H.~Zhang, B.~Chen, Z.~Meng, and X.~Yuan, ``Birnat:
  Bidirectional recurrent neural networks with adversarial training for video
  snapshot compressive imaging,'' in \emph{European Conference on Computer
  Vision}.\hskip 1em plus 0.5em minus 0.4em\relax Springer, 2020, pp. 258--275.

\bibitem{cheng2021memory}
Z.~Cheng, B.~Chen, G.~Liu, H.~Zhang, R.~Lu, Z.~Wang, and X.~Yuan,
  ``Memory-efficient network for large-scale video compressive sensing,'' in
  \emph{Proceedings of the IEEE/CVF Conference on Computer Vision and Pattern
  Recognition}, 2021, pp. 16\,246--16\,255.

\bibitem{ma2019deep}
J.~Ma, X.-Y. Liu, Z.~Shou, and X.~Yuan, ``Deep tensor admm-net for snapshot
  compressive imaging,'' in \emph{Proceedings of the IEEE/CVF International
  Conference on Computer Vision}, 2019, pp. 10\,223--10\,232.

\bibitem{wang2022spatial}
L.~Wang, M.~Cao, Y.~Zhong, and X.~Yuan, ``Spatial-temporal transformer for
  video snapshot compressive imaging,'' \emph{IEEE Transactions on Pattern
  Analysis and Machine Intelligence}, 2022.

\bibitem{wang2023efficientsci}
L.~Wang, M.~Cao, and X.~Yuan, ``Efficientsci: Densely connected network with
  space-time factorization for large-scale video snapshot compressive
  imaging,'' in \emph{Proceedings of the IEEE/CVF Conference on Computer Vision
  and Pattern Recognition}, 2023, pp. 18\,477--18\,486.

\bibitem{cao2024hybrid}
M.~Cao, L.~Wang, M.~Zhu, and X.~Yuan, ``Hybrid cnn-transformer architecture for
  efficient large-scale video snapshot compressive imaging,''
  \emph{International Journal of Computer Vision}, pp. 1--20, 2024.

\bibitem{mildenhall2021nerf}
B.~Mildenhall, P.~P. Srinivasan, M.~Tancik, J.~T. Barron, R.~Ramamoorthi, and
  R.~Ng, ``Nerf: Representing scenes as neural radiance fields for view
  synthesis,'' \emph{Communications of the ACM}, vol.~65, no.~1, pp. 99--106,
  2021.

\bibitem{schonberger2016structure}
J.~L. Schonberger and J.-M. Frahm, ``Structure-from-motion revisited,'' in
  \emph{Proceedings of the IEEE conference on computer vision and pattern
  recognition}, 2016, pp. 4104--4113.

\bibitem{kerbl20233d}
B.~Kerbl, G.~Kopanas, T.~Leimk{\"u}hler, and G.~Drettakis, ``3d gaussian
  splatting for real-time radiance field rendering.'' \emph{ACM Trans. Graph.},
  vol.~42, no.~4, pp. 139--1, 2023.

\bibitem{kheradmand20243d}
S.~Kheradmand, D.~Rebain, G.~Sharma, W.~Sun, J.~Tseng, H.~Isack, A.~Kar,
  A.~Tagliasacchi, and K.~M. Yi, ``3d gaussian splatting as markov chain monte
  carlo,'' \emph{arXiv preprint arXiv:2404.09591}, 2024.

\bibitem{li2024scinerf}
Y.~Li, X.~Wang, P.~Wang, X.~Yuan, and P.~Liu, ``{SCINeRF: Neural Radiance
  Fields from a Snapshot Compressive Image},'' in \emph{Proceedings of the
  IEEE/CVF Conference on Computer Vision and Pattern Recognition}, 2024, pp.
  10\,542--10\,552.

\bibitem{yang2020shearlet}
P.~Yang, L.~Kong, X.-Y. Liu, X.~Yuan, and G.~Chen, ``Shearlet enhanced snapshot
  compressive imaging,'' \emph{IEEE Transactions on Image Processing}, vol.~29,
  pp. 6466--6481, 2020.

\bibitem{boyd2011distributed}
S.~Boyd, N.~Parikh, E.~Chu, B.~Peleato, J.~Eckstein \emph{et~al.},
  ``Distributed optimization and statistical learning via the alternating
  direction method of multipliers,'' \emph{Foundations and
  Trends{\textregistered} in Machine learning}, vol.~3, no.~1, pp. 1--122,
  2011.

\bibitem{ronneberger2015u}
O.~Ronneberger, P.~Fischer, and T.~Brox, ``U-net: Convolutional networks for
  biomedical image segmentation,'' in \emph{Medical Image Computing and
  Computer-Assisted Intervention--MICCAI 2015: 18th International Conference,
  Munich, Germany, October 5-9, 2015, Proceedings, Part III 18}.\hskip 1em plus
  0.5em minus 0.4em\relax Springer, 2015, pp. 234--241.

\bibitem{NIPS2014_5ca3e9b1}
I.~Goodfellow, J.~Pouget-Abadie, M.~Mirza, B.~Xu, D.~Warde-Farley, S.~Ozair,
  A.~Courville, and Y.~Bengio, ``Generative adversarial nets,'' \emph{Advances
  in neural information processing systems}, vol.~27, 2014.

\bibitem{johnson2016perceptual}
J.~Johnson, A.~Alahi, and L.~Fei-Fei, ``Perceptual losses for real-time style
  transfer and super-resolution,'' in \emph{Computer Vision--ECCV 2016: 14th
  European Conference, Amsterdam, The Netherlands, October 11-14, 2016,
  Proceedings, Part II 14}.\hskip 1em plus 0.5em minus 0.4em\relax Springer,
  2016, pp. 694--711.

\bibitem{miao2019net}
X.~Miao, X.~Yuan, Y.~Pu, and V.~Athitsos, ``l-net: Reconstruct hyperspectral
  images from a snapshot measurement,'' in \emph{Proceedings of the IEEE/CVF
  International Conference on Computer Vision}, 2019, pp. 4059--4069.

\bibitem{he2016deep}
K.~He, X.~Zhang, S.~Ren, and J.~Sun, ``Deep residual learning for image
  recognition,'' in \emph{Proceedings of the IEEE conference on computer vision
  and pattern recognition}, 2016, pp. 770--778.

\bibitem{yuan2020plug}
X.~Yuan, Y.~Liu, J.~Suo, and Q.~Dai, ``Plug-and-play algorithms for large-scale
  snapshot compressive imaging,'' in \emph{Proceedings of the IEEE/CVF
  Conference on Computer Vision and Pattern Recognition}, 2020, pp. 1447--1457.

\bibitem{yuan2021plug}
X.~Yuan, Y.~Liu, J.~Suo, F.~Durand, and Q.~Dai, ``Plug-and-play algorithms for
  video snapshot compressive imaging,'' \emph{IEEE Transactions on Pattern
  Analysis and Machine Intelligence}, vol.~44, no.~10, pp. 7093--7111, 2021.

\bibitem{vaswani2017attention}
A.~Vaswani, N.~Shazeer, N.~Parmar, J.~Uszkoreit, L.~Jones, A.~N. Gomez,
  {\L}.~Kaiser, and I.~Polosukhin, ``Attention is all you need,''
  \emph{Advances in neural information processing systems}, vol.~30, 2017.

\bibitem{gu2021stylenerf}
J.~Gu, L.~Liu, P.~Wang, and C.~Theobalt, ``Stylenerf: A style-based 3d-aware
  generator for high-resolution image synthesis,'' 2021.

\bibitem{xiangli2022bungeenerf}
Y.~Xiangli, L.~Xu, X.~Pan, N.~Zhao, A.~Rao, C.~Theobalt, B.~Dai, and D.~Lin,
  ``Bungeenerf: Progressive neural radiance field for extreme multi-scale scene
  rendering,'' in \emph{European conference on computer vision}.\hskip 1em plus
  0.5em minus 0.4em\relax Springer, 2022, pp. 106--122.

\bibitem{turki2022mega}
H.~Turki, D.~Ramanan, and M.~Satyanarayanan, ``Mega-nerf: Scalable construction
  of large-scale nerfs for virtual fly-throughs,'' in \emph{Proceedings of the
  IEEE/CVF Conference on Computer Vision and Pattern Recognition}, 2022, pp.
  12\,922--12\,931.

\bibitem{tancik2020fourier}
M.~Tancik, P.~Srinivasan, B.~Mildenhall, S.~Fridovich-Keil, N.~Raghavan,
  U.~Singhal, R.~Ramamoorthi, J.~Barron, and R.~Ng, ``Fourier features let
  networks learn high frequency functions in low dimensional domains,''
  \emph{Advances in Neural Information Processing Systems}, vol.~33, pp.
  7537--7547, 2020.

\bibitem{athar2022rignerf}
S.~Athar, Z.~Xu, K.~Sunkavalli, E.~Shechtman, and Z.~Shu, ``Rignerf: Fully
  controllable neural 3d portraits,'' in \emph{Proceedings of the IEEE/CVF
  conference on Computer Vision and Pattern Recognition}, 2022, pp.
  20\,364--20\,373.

\bibitem{gafni2021dynamic}
G.~Gafni, J.~Thies, M.~Zollhofer, and M.~Nie{\ss}ner, ``Dynamic neural radiance
  fields for monocular 4d facial avatar reconstruction,'' in \emph{Proceedings
  of the IEEE/CVF Conference on Computer Vision and Pattern Recognition}, 2021,
  pp. 8649--8658.

\bibitem{peng2021animatable}
S.~Peng, J.~Dong, Q.~Wang, S.~Zhang, Q.~Shuai, X.~Zhou, and H.~Bao,
  ``Animatable neural radiance fields for modeling dynamic human bodies,'' in
  \emph{Proceedings of the IEEE/CVF International Conference on Computer
  Vision}, 2021, pp. 14\,314--14\,323.

\bibitem{peng2021neural}
S.~Peng, Y.~Zhang, Y.~Xu, Q.~Wang, Q.~Shuai, H.~Bao, and X.~Zhou, ``Neural
  body: Implicit neural representations with structured latent codes for novel
  view synthesis of dynamic humans,'' in \emph{Proceedings of the IEEE/CVF
  Conference on Computer Vision and Pattern Recognition}, 2021, pp. 9054--9063.

\bibitem{Huang_2022_CVPR}
X.~Huang, Q.~Zhang, Y.~Feng, H.~Li, X.~Wang, and Q.~Wang, ``Hdr-nerf: High
  dynamic range neural radiance fields,'' in \emph{Proceedings of the IEEE/CVF
  Conference on Computer Vision and Pattern Recognition (CVPR)}, June 2022, pp.
  18\,398--18\,408.

\bibitem{yang2021learning}
B.~Yang, Y.~Zhang, Y.~Xu, Y.~Li, H.~Zhou, H.~Bao, G.~Zhang, and Z.~Cui,
  ``Learning object-compositional neural radiance field for editable scene
  rendering,'' in \emph{Proceedings of the IEEE/CVF International Conference on
  Computer Vision}, 2021, pp. 13\,779--13\,788.

\bibitem{yuan2022nerf}
Y.-J. Yuan, Y.-T. Sun, Y.-K. Lai, Y.~Ma, R.~Jia, and L.~Gao, ``Nerf-editing:
  geometry editing of neural radiance fields,'' in \emph{Proceedings of the
  IEEE/CVF Conference on Computer Vision and Pattern Recognition}, 2022, pp.
  18\,353--18\,364.

\bibitem{weder2023removing}
S.~Weder, G.~Garcia-Hernando, A.~Monszpart, M.~Pollefeys, G.~J. Brostow,
  M.~Firman, and S.~Vicente, ``Removing objects from neural radiance fields,''
  in \emph{Proceedings of the IEEE/CVF Conference on Computer Vision and
  Pattern Recognition}, 2023, pp. 16\,528--16\,538.

\bibitem{wei2023clutter}
F.~Wei, T.~Funkhouser, and S.~Rusinkiewicz, ``Clutter detection and removal in
  3d scenes with view-consistent inpainting,'' in \emph{Proceedings of the
  IEEE/CVF International Conference on Computer Vision}, 2023, pp.
  18\,131--18\,141.

\bibitem{bi2020neural}
S.~Bi, Z.~Xu, P.~Srinivasan, B.~Mildenhall, K.~Sunkavalli, M.~Ha{\v{s}}an,
  Y.~Hold-Geoffroy, D.~Kriegman, and R.~Ramamoorthi, ``Neural reflectance
  fields for appearance acquisition,'' \emph{arXiv preprint arXiv:2008.03824},
  2020.

\bibitem{boss2021nerd}
M.~Boss, R.~Braun, V.~Jampani, J.~T. Barron, C.~Liu, and H.~Lensch, ``Nerd:
  Neural reflectance decomposition from image collections,'' in
  \emph{Proceedings of the IEEE/CVF International Conference on Computer
  Vision}, 2021, pp. 12\,684--12\,694.

\bibitem{wang2021nerf}
Z.~Wang, S.~Wu, W.~Xie, M.~Chen, and V.~A. Prisacariu, ``Nerf--: Neural
  radiance fields without known camera parameters,'' \emph{arXiv preprint
  arXiv:2102.07064}, 2021.

\bibitem{jeong2021self}
Y.~Jeong, S.~Ahn, C.~Choy, A.~Anandkumar, M.~Cho, and J.~Park,
  ``Self-calibrating neural radiance fields,'' in \emph{Proceedings of the
  IEEE/CVF International Conference on Computer Vision}, 2021, pp. 5846--5854.

\bibitem{sucar2021imap}
E.~Sucar, S.~Liu, J.~Ortiz, and A.~J. Davison, ``imap: Implicit mapping and
  positioning in real-time,'' in \emph{Proceedings of the IEEE/CVF
  International Conference on Computer Vision}, 2021, pp. 6229--6238.

\bibitem{meng2021gnerf}
Q.~Meng, A.~Chen, H.~Luo, M.~Wu, H.~Su, L.~Xu, X.~He, and J.~Yu, ``Gnerf:
  Gan-based neural radiance field without posed camera,'' in \emph{Proceedings
  of the IEEE/CVF International Conference on Computer Vision}, 2021, pp.
  6351--6361.

\bibitem{lin2021barf}
C.-H. Lin, W.-C. Ma, A.~Torralba, and S.~Lucey, ``Barf: Bundle-adjusting neural
  radiance fields,'' in \emph{Proceedings of the IEEE/CVF International
  Conference on Computer Vision}, 2021, pp. 5741--5751.

\bibitem{Wang_2023_CVPR}
P.~Wang, L.~Zhao, R.~Ma, and P.~Liu, ``Bad-nerf: Bundle adjusted deblur neural
  radiance fields,'' in \emph{Proceedings of the IEEE/CVF Conference on
  Computer Vision and Pattern Recognition (CVPR)}, June 2023, pp. 4170--4179.

\bibitem{chen2022tensorf}
A.~Chen, Z.~Xu, A.~Geiger, J.~Yu, and H.~Su, ``{TensoRF: Tensorial Radiance
  Fields},'' in \emph{ECCV}.\hskip 1em plus 0.5em minus 0.4em\relax Springer,
  2022, pp. 333--350.

\bibitem{fridovich2022plenoxels}
S.~Fridovich-Keil, A.~Yu, M.~Tancik, Q.~Chen, B.~Recht, and A.~Kanazawa,
  ``{Plenoxels: Radiance Fields without Neural Networks},'' in \emph{CVPR},
  2022, pp. 5501--5510.

\bibitem{cao2023hexplane}
A.~Cao and J.~Johnson, ``Hexplane: A fast representation for dynamic scenes,''
  in \emph{Proceedings of the IEEE/CVF Conference on Computer Vision and
  Pattern Recognition}, 2023, pp. 130--141.

\bibitem{mueller2022instant}
\BIBentryALTinterwordspacing
T.~M\"uller, A.~Evans, C.~Schied, and A.~Keller, ``{Instant Neural Graphics
  Primitives with a Multiresolution Hash Encoding},'' vol.~41, no.~4, pp.
  102:1--102:15, Jul. 2022. [Online]. Available:
  \url{https://doi.org/10.1145/3528223.3530127}
\BIBentrySTDinterwordspacing

\bibitem{fan2024instantsplat}
Z.~Fan, W.~Cong, K.~Wen, K.~Wang, J.~Zhang, X.~Ding, D.~Xu, B.~Ivanovic,
  M.~Pavone, G.~Pavlakos \emph{et~al.}, ``Instantsplat: Unbounded sparse-view
  pose-free gaussian splatting in 40 seconds,'' \emph{arXiv preprint
  arXiv:2403.20309}, 2024.

\bibitem{wang2024dust3r}
S.~Wang, V.~Leroy, Y.~Cabon, B.~Chidlovskii, and J.~Revaud, ``Dust3r: Geometric
  3d vision made easy,'' in \emph{Proceedings of the IEEE/CVF Conference on
  Computer Vision and Pattern Recognition}, 2024, pp. 20\,697--20\,709.

\bibitem{wang2024vggsfm}
J.~Wang, N.~Karaev, C.~Rupprecht, and D.~Novotny, ``Vggsfm: Visual geometry
  grounded deep structure from motion,'' in \emph{Proceedings of the IEEE/CVF
  Conference on Computer Vision and Pattern Recognition}, 2024, pp.
  21\,686--21\,697.

\bibitem{HH}
\BIBentryALTinterwordspacing
H.~Heo, ``Radiance fields from vggsfm and mast3r, and their comparison,'' 2024.
  [Online]. Available:
  \url{https://github.com/hwanhuh/Radiance-Fields-from-VGGSfM-Mast3r}
\BIBentrySTDinterwordspacing

\bibitem{bulo2024revising}
S.~R. Bul{\`o}, L.~Porzi, and P.~Kontschieder, ``Revising densification in
  gaussian splatting,'' \emph{arXiv preprint arXiv:2404.06109}, 2024.

\bibitem{zhang2024pixel}
Z.~Zhang, W.~Hu, Y.~Lao, T.~He, and H.~Zhao, ``Pixel-gs: Density control with
  pixel-aware gradient for 3d gaussian splatting,'' \emph{arXiv preprint
  arXiv:2403.15530}, 2024.

\bibitem{ye2024absgs}
Z.~Ye, W.~Li, S.~Liu, P.~Qiao, and Y.~Dou, ``Absgs: Recovering fine details in
  3d gaussian splatting,'' in \emph{ACM Multimedia 2024}, 2024.

\bibitem{wang2004image}
Z.~Wang, A.~C. Bovik, H.~R. Sheikh, and E.~P. Simoncelli, ``Image quality
  assessment: From error visibility to structural similarity,'' \emph{IEEE
  Transactions on Image Processing}, vol.~13, no.~4, pp. 600--612, 2004.

\bibitem{image_loss}
H.~Zhao, O.~Gallo, I.~Frosio, and J.~Kautz, ``Loss functions for image
  restoration with neural networks,'' \emph{IEEE Transactions on Computational
  Imaging}, vol.~3, no.~1, pp. 47--57, 2017.

\bibitem{mildenhall2019local}
B.~Mildenhall, P.~P. Srinivasan, R.~Ortiz-Cayon, N.~K. Kalantari,
  R.~Ramamoorthi, R.~Ng, and A.~Kar, ``Local light field fusion: Practical view
  synthesis with prescriptive sampling guidelines,'' \emph{ACM Transactions on
  Graphics (TOG)}, vol.~38, no.~4, pp. 1--14, 2019.

\bibitem{ma2022deblur}
L.~Ma, X.~Li, J.~Liao, Q.~Zhang, X.~Wang, J.~Wang, and P.~V. Sander,
  ``Deblur-nerf: Neural radiance fields from blurry images,'' in
  \emph{Proceedings of the IEEE/CVF Conference on Computer Vision and Pattern
  Recognition}, 2022, pp. 12\,861--12\,870.

\bibitem{zhang2018unreasonable}
R.~Zhang, P.~Isola, A.~A. Efros, E.~Shechtman, and O.~Wang, ``The unreasonable
  effectiveness of deep features as a perceptual metric,'' in \emph{Proceedings
  of the IEEE conference on computer vision and pattern recognition (CVPR)},
  2018, pp. 586--595.

\bibitem{lin2020nerfpytorch}
L.~Yen-Chen, ``Nerf-pytorch,''
  \url{https://github.com/yenchenlin/nerf-pytorch/}, 2020.

\bibitem{kingma2017adam}
D.~P. Kingma, ``Adam: A method for stochastic optimization,'' \emph{arXiv
  preprint arXiv:1412.6980}, 2014.

\bibitem{andre2022sqr}
J.~Andr{\'e}, F.~Strati, and A.~Klimovic, ``Exploring learning rate scaling
  rules for distributed ml training on transient resources,'' in
  \emph{Proceedings of the 3rd International Workshop on Distributed Machine
  Learning}.\hskip 1em plus 0.5em minus 0.4em\relax ACM, 2022, pp. 1--8.

\end{thebibliography}



\begin{IEEEbiographynophoto}{Yunhao Li}
received his B.Eng. degree in Information Engineering from Xi'an Jiaotong University, Xi'an, China, in 2019, and his M.S. degree in Computer Engineering from Northwestern University, Evanston, IL, USA, in 2021. He is currently pursuing a Ph.D. in Computer Science and Engineering at Zhejiang University, jointly with Westlake University. His research interests include computer vision, 3D reconstruction, and computational imaging.
\end{IEEEbiographynophoto}

\begin{IEEEbiographynophoto}{Xiang Liu}
received his B.Eng. degree in Mechanical Design, Manufacture, and Automation from Hunan University, Changsha, China, in 2020, and his M.Sc. degree in Robotics, Systems, and Control from ETH Zurich, Switzerland, in 2024. He is currently a research assistant at Westlake University. His research interests span 3D vision, egocentric vision, and robotics.
\end{IEEEbiographynophoto}

\begin{IEEEbiographynophoto}{Xiaodong Wang}
received his B.S. degree from Changchun University of Science and Technology, Changchun, China, in 2019, and his M.S. degree from Southern University of Science and Technology, Shenzhen, China, in 2022. He is currently a third-year Ph.D. candidate at Westlake University, Hangzhou, China. His research interests include computational imaging, low-level vision tasks, and 3D reconstruction.
\end{IEEEbiographynophoto}

\begin{IEEEbiographynophoto}{Xin Yuan}
(Senior Member, IEEE) received the B.Eng. and M.Eng. degrees from Xidian University, China, in 2007 and 2009, respectively, and the Ph.D. degree from Hong Kong Polytechnic University, Hong Kong, in 2012. He is currently an Associate Professor at Westlake University, Hangzhou, China. From 2015 to 2021, he served as a lead researcher in video analysis and coding at Bell Labs, Murray Hill, NJ, USA. Prior to that, he was a Postdoctoral Associate in the Department of Electrical and Computer Engineering at Duke University from 2012 to 2015.
His research interests include signal processing, computational imaging, and machine learning. He has been an Associate Editor for Pattern Recognition since 2019, International Journal of Pattern Recognition and Artificial Intelligence since 2020, a Topic Editor for Chinese Optics Letters since 2021, and a Receiving Editor for Optics and Laser Technology since 2024. He led the special issue on "Deep Learning for High-Dimensional Sensing" in the IEEE Journal of Selected Topics in Signal Processing in 2022.
\end{IEEEbiographynophoto}

\begin{IEEEbiographynophoto}{Peidong Liu}
received the B.E. and M.E. degrees in Electrical Engineering from the National University of Singapore, Singapore, in 2012 and 2015, respectively, and the Ph.D. degree in Computer Science from ETH Zurich, Zurich, Switzerland, in 2021. He is currently an Assistant Professor in the School of Engineering at Westlake University, Hangzhou, China. His research interests include 3D computer vision and robotics.
\end{IEEEbiographynophoto}

\end{document}